\documentclass[runningheads]{llncs}

\usepackage{eccv}

\usepackage{eccvabbrv}

\usepackage{graphicx}
\usepackage{booktabs}
\usepackage[accsupp]{axessibility}  % Improves PDF readability for those with disabilities.
\usepackage{multirow}   % for sec/5_exp.tex tables
\usepackage{makecell}   % for sec/5_exp.tex tables
\usepackage{placeins}   % \FloatBarrier: keep figures within their section
\usepackage{wrapfig}    % figure and text side by side (half column)
\usepackage{caption}    % \captionof for side-by-side tables
\usepackage[most]{tcolorbox} % rounded highlight boxes

\usepackage[pagebackref,breaklinks,colorlinks,citecolor=eccvblue]{hyperref}
\usepackage{orcidlink}

\begin{document}

\title{Goal-oriented Navigation Instruction Generation with Tour Video Priors} 

% \titlerunning{Navigation Instruction Generation with Tour Video Priors}

\author{
    Fangdi Li\inst{1,2*}\and
    Juncheng Liao\inst{1,2*}\and
    Changxu Cheng\inst{1*}\and
    Jiazhi Wang\inst{1*} \and \\
    Senda Chen\inst{1,3} \and
    Tao Wang\inst{1}$^{\dagger}$ \and
    Wuyue Zhao\inst{1}
}

\authorrunning{F.~Li et al.}

\institute{
    Uni-Ubi AI, Hangzhou, China \and
    Zhejiang University, Hangzhou, China \and
    Tongji University, Shanghai, China
}
\maketitle

\renewcommand{\thefootnote}{}\footnotetext{$^{\!*}$~Equal contribution.}
\renewcommand{\thefootnote}{}\footnotetext{$^{\!\dagger}$~Corresponding author. Email: wangtaomarvel@gmail.com}

\begin{abstract}
Navigation Instruction Generation (NIG) aims to produce step-by-step natural language instructions for navigation guidance.
% "reversed task" -> inverse task
Existing studies primarily treat NIG as an auxiliary task for vision-and-language navigation (VLN), focusing on data augmentation or multi-task learning.
However, generating navigation instructions from compact environmental priors requires meticulous spatial reasoning, especially when the target route does not simply follow the demonstrated tour, and remains challenging for current multimodal models.
In this work, we introduce VideoNIG, a goal-oriented video-grounded NIG task that generates navigation instructions from ego-centric tour videos, an initial observation, and a textual or visual goal, without relying on intermediate representations such as graphs and maps.
% multimodal prompts -> instructional samples
We instantiate VideoNIG in a controlled simulator benchmark with 60K tour videos across continuous indoor environments and 37K multimodal prompts with progressive difficulty levels.
% inverse dynamics modeling 
% To address this task, we propose a two-stage Curriculum Learning framework that decomposes the learning into inverse dynamics modeling and progressively complex path reasoning.
% Basic motion perception
We further introduce a diagnostic evaluation protocol that combines text similarity, choice-based spatial consistency tests, and downstream navigation execution.
To address this task, we propose a two-stage Curriculum Learning framework that decomposes the learning into foundational motion perception and long-horizon navigation reasoning.
% atomic
% Specifically, we first employ Action Warmup to establish robust spatial grounding between actions and visual changes, followed by Complexity Progression training using trajectories with increasingly exploratory behaviors and navigational difficulty
% followed by Complexity Progression using trajectories with increasing exploratory difficulty.
Specifically, we first employ Action Warmup for spatial action-view alignment, followed by Complexity Progression using trajectories with increasing exploratory difficulty.
% Specifically, we first employ Action Warmup for spatial action-view alignment, followed by Complexity Progression training using tour videos with decreasing overlap with optimal trajectories.
Extensive experiments show that existing MLLMs struggle with VideoNIG, while our approach significantly improves instruction quality across complementary diagnostic metrics.
% Extensive experiments show that while existing MLLMs struggle with the VideoNIG task, our approach significantly boosts both accuracy and robustness.
Finally, integrating VideoNIG-generated instructions with a VLN agent demonstrates the executability of this task formulation for end-to-end navigation.

% Navigation Instruction Generation (NIG) is the task of generating step-by-step natural language instructions to guide navigation.
% Most previous studies regarded NIG as an auxiliary task to vision-and-language navigation (VLN), primarily for data augmentation or multi-task learning.
% However, fine-grained navigation instructions require meticulous reasoning, making them less user-friendly in persistent navigation settings.
% In this work, we propose a novel task named VideoNIG -- to generate goal-oriented navigation instructions using naive environmental priors, \ie , tour videos.
% Specifically, we instantiate VideoNIG with 60K tour videos from continuous environments in the Habitat simulator, and 37K discriminative multimodal prompts.

% key words

\end{abstract}

\section{Introduction}
\label{sec:intro}
% introduce the key concepts
Vision-and-Language Navigation (VLN) has become an important and widely studied topic in the embodied intelligence community~\cite{anderson2018r2r,ku2020rxr,krantz2020vlnce,chen2022think,zheng2024navillm,hong2025general}.
With the powerful reasoning capabilities of language models, language plays a key role in improving navigation generalization and facilitating human-robot interaction~\cite{zhang2024vision,zhou2024navgpt,han2025dialnav}.
As a result, Navigation Instruction Generation (NIG)~\cite{fried2018speaker,tan2019learning,wang2023lana,zheng2024navillm,cui2025generating} has also been actively explored as a complementary direction that aims to generate natural language guidance from observations.

\begin{figure}[t]
    \centering
    \includegraphics[width=1\linewidth]{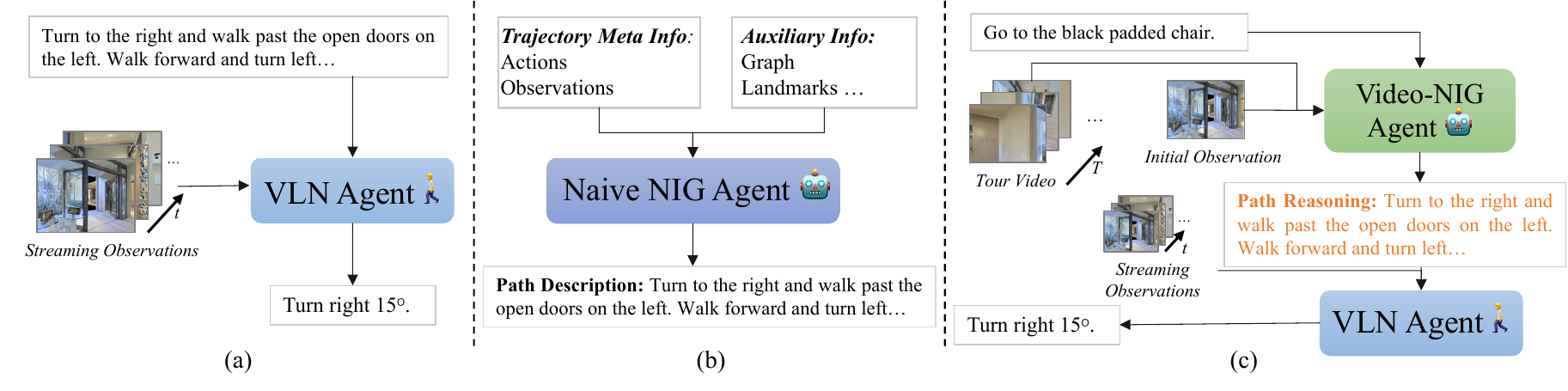}
    \vspace{-2mm}
    \caption{
        The relationship between NIG and VLN.
        (a) The input to a VLN agent is typically fine-grained textual guidance, which demands extensive reasoning from users and is therefore not user-friendly.
        (b) Most previous works formulate NIG as a trajectory-captioning task, optionally with auxiliary heuristic environmental information.
        (c) Our VideoNIG agent requires only simple goal descriptions from users, together with single-view ego-centric tour videos and initial observations, to generate global navigation instructions.
        This relieves human users from the reasoning effort required to prepare inputs for VLN agents.
    }
    \vspace{-4mm}
    \label{fig:intro_rel}
\end{figure}

% previous NIG&VLN; 引出NIG
\noindent Most existing works treat NIG merely as trajectory captioning (\cref{fig:intro_rel}(b)), using it to augment VLN training data or facilitate multi-task learning~\cite{fried2018speaker,tan2019learning,wang2023scaling,zheng2024navillm,wang2023lana,wang2022counterfactual}.
Although VLN has made substantial progress, the fine-grained navigation instructions provided as input are crucial to navigation performance. An effective navigation instruction should allow a navigator to reach the goal without ambiguity in diverse environments when followed strictly~\cite{ku2020rxr}.
This demands meticulous reasoning from users, often yielding suboptimal or ambiguous instructions, which is not user-friendly, as shown in \cref{fig:intro_rel}(a).
Scenario-oriented object navigation~\cite{zhu2021soon} is thus introduced to simplify human input into object-level descriptions, with the starting point unconstrained.
In this setting, the agent explores the environment toward the target, optionally leveraging explicit environment representations like frontier maps~\cite{yokoyama2024vlfm}.
Considering both goal-oriented navigation and fine-grained instructions in VLN, this raises the question: can optimal navigation instructions be automatically generated for global planning, given a goal description and environmental priors?
% TODO: compress this part
GoViG~\cite{wu2025govig} generates instructions solely from start-goal observations to bypass the limitations of explicit environmental priors. 
Despite its adaptability, GoViG is prone to hallucinations over long distances. 
Furthermore, its zero-shot instructions often remain suboptimal, exhibiting ambiguities or inefficient path planning relative to shortest-path trajectories in persistent indoor settings.
% A recent work, GoViG~\cite{wu2025govig}, generates navigation instructions using only initial and goal observations without environmental priors.
% It is argued that previous explicit environment representations are either limited to structured scenarios or discarding some essential information from observations.
% Despite the high adapability to unseen environments, GoViG is likely to hallucinate once the start and goal points are far apart. 
% Besides, the zero-shot generated navigation instruction is prone to be suboptimal.
% For example, the instructions may contain ambiguities, and the planned path can be significantly longer than the shortest one, when grounded in real environments.
% 没有必要一直做 zero-shot exploration，应该要有环境记忆
Therefore, environmental priors are crucial for goal-oriented NIG, and zero-shot navigation should not be overemphasized, as embodied agents often operate in persistent environments~\cite{hong2025general}.

% 引出 tour video priors
\noindent Recent advances in video-based visual spatial intelligence demonstrate the potential of vision-language models for understanding the physical world~\cite{yang2025thinking,zhou2025vlm4d,huang2025mllms}.
Learning actions from videos has also been explored in robot manipulation~\cite{chen2025moto}.
% introduce VideoNIG
This motivates us to use ego-centric tour videos as compact video-native environmental priors, which provide dense and temporally ordered spatial cues while avoiding explicit representations such as landmarks~\cite{wang2022less}, BEV features~\cite{fan2024navigation}, and topological maps~\cite{an2024etpnav}.
% 解释tour video
\noindent Here, a tour video refers to a recording captured by humans or embodied agents that covers a wide range of the environment with rich visual and spatial information and is usually associated with the target route.
As a result, we propose VideoNIG, a novel task of goal-oriented navigation instruction generation that is conditioned on ego-centric tour videos, initial observations, and goal descriptions.
\cref{fig:intro_rel}(c) illustrates the concept of VideoNIG, in which the NIG agent performs global planning autonomously while requiring from the user only a simple goal description.
% VideoNIG VS GoViG VS trajectory_description
Endpoint-only formulations such as GoViG~\cite{wu2025govig} and trajectory-captioning NIG can be viewed as neighboring settings along the amount of available environmental prior: they use either sparse endpoint observations or a tour that exactly follows the target route, whereas VideoNIG studies richer tour-video priors that may include additional observations beyond the optimal path.

% 介绍数据集
% \noindent To support the task, we construct the VideoNIG dataset through an elaborate data pipeline, comprising 60K tour videos with three different construction methods, 20K gold route videos marking the optimal paths, and 37K multimodal discriminative goal descriptions.
\noindent To instantiate the task, we build a controlled VideoNIG benchmark comprising 60K tour videos with three different complexity levels, 20K Gold Route videos marking the optimal paths, and 37K multimodal discriminative goal descriptions.
While ensuring coverage of the optimal path's viewpoints, the tours are further diversified by randomly extending start and end points, introducing random look-around behaviors along the route, and shifting waypoints to other local reachable locations. These settings provide a graded experimental substrate for analyzing video-grounded spatial reasoning.
% our method: two-stage Curriculum Learning method (GRPO/MPO-training)
Meanwhile, we propose a two-stage Curriculum Learning framework, consisting of Action Warmup and Complexity Progression to progressively enhance the visual-spatial reasoning capabilities of multimodal large language models (MLLMs). 
% evaluation metric: text similarity / choice benchmark / navigation execution
% The evaluation metric combines classical text similarity measures with a vision-language model\-based scoring mechanism, providing a faithful assessment of content accuracy.
% We adopt two complementary evaluation perspectives: classical text similarity and Choice Evaluation.
We adopt three complementary evaluation perspectives: classical text similarity metrics, Choice Evaluation and Navigation Execution.
% VLN Agent
We integrate VideoNIG with an MLLM-based VLN agent to assess downstream navigation performance in an end-to-end setting from tour videos and goal descriptions.
% To comprehensively assess instruction quality, we evaluate our model from three complementary evaluation perspectives: (1) standard text similarity metrics, Rouge-L and SPICE, for lexical and semantic alignment; (2) two newly proposed spatially diagnostic benchmarks that explicitly probe directional consistency; and (3) downstream VLN agent performance, which evaluates the executability of generated instructions in embodied settings.
% 评估MLLM的轨迹描述能力、定位能力、空间推理能力
% Performance consistently degrades with increasing video complexity.
Our experiments reveal that performance consistently degrades as video complexity increases, highlighting VideoNIG as a challenging yet promising task.
% Our evaluation results suggest that performance consistently degrades with increasing video complexity and VideoNIG is a challenging but promising task.
% Existing state-of-the-art multimodal LLMs (MLLMs) perform well when the tour degenerates into the (optimal) gold route, but their performance drops sharply once the tour becomes more complex and natural, rather than a single uninterrupted path exactly leading to the goal.
% Even their improvements on visual spatial understanding benchmarks do not transfer well to our task.
% 这个不太好啊，和后面的线性分析冲突了
% Existing state-of-the-art multimodal LLMs (MLLMs) perform poorly on VideoNIG, suggesting that their improvements on visual spatial understanding benchmarks do not transfer well to our task.
Existing state-of-the-art MLLMs perform poorly on VideoNIG, suggesting that their spatial reasoning abilities, although improved on existing benchmarks, are not effectively activated by this task.
Our analysis suggests this limitation primarily manifests as semantic fragmentation in the generated instructions: models struggle to maintain persistent spatial grounding throughout long-horizon navigation videos, which in turn misleads the downstream agent.
% This limitation likely arises from difficulties in maintaining consistent localization and spatial grounding throughout long-horizon navigation videos.
% Existing state-of-the-art multimodal LLMs (MLLMs) perform poorly on VideoNIG, suggesting that their improvements on visual spatial understanding benchmarks do not transfer well to our task.
% This reflects that the localization and spatial reasoning abilities of these models are not effectively activated in practical scenarios.
% This indicates that the localization and spatial reasoning abilities of these models are not effectively activated in practical navigation scenarios.
% Further analysis shows that this limitation mainly stems from semantic fragmentation in generated instructions: the models struggle to maintain consistent spatial grounding across complex scenes, which in turn misleads the downstream navigation agent.
% Further analysis reveals that this is primarily due to the semantic fragmentation in the generated instructions; the MLLM struggles to maintain persistent spatial grounding in complex scenes, which in turn misleads the navigation agent.
% Surprisingly, when the tour video corresponds to the gold route, the generated instructions yield a higher navigation success rate than the original human annotations.

\noindent Succinctly, our contributions are threefold:
(1) we define \textbf{VideoNIG}, a goal-oriented video-grounded NIG task that converts tour-video environmental priors, initial observations, and multimodal goals into executable navigation instructions.
(2) we introduce a diagnostic evaluation protocol that combines Choice Evaluation and Navigation Execution to assess spatial consistency and downstream executability beyond text similarity.
(3) we provide a two-stage Curriculum Learning reference method and a controlled benchmark instantiation, revealing persistent limitations of current MLLMs on long-horizon video-grounded navigation reasoning.
% (1) we introduce \textbf{VideoNIG}, a novel task for user-friendly and autonomous navigation where only tour videos and high-level goal descriptions are required;
% (2) we instantiate \textbf{VideoNIG} with progressively challenging tours and multimodal discriminative goal descriptions in a controlled benchmark;
% (3) we propose a two-stage Curriculum Learning framework that activates the visual-spatial capabilities of MLLMs, significantly improving performance on the VideoNIG task.

% 轨迹长度的难度差异

% 目标点表示方式的难度差异

% 评估视频空间智能在真实场景的效果

% thinking的影响

% generate "optimal instructions"

% Efficient Evaluation. Basic elements (eg, directions), effects on downstream VLN models

\FloatBarrier
\section{Related Works}
\label{sec:related}

\noindent \textbf{Vision-and-Language Navigation} (VLN) requires an embodied agent to execute navigation actions based on natural language instructions and visual observations~\cite{anderson2018r2r,ku2020rxr,krantz2020vlnce}.
Typical instructions are detailed and step-by-step, enabling zero-shot navigation in unseen environments.
% map-free
Map-free approaches such as Navid~\cite{zhang2024navid} and NaviLLM~\cite{zheng2024navillm} leverage large vision-language models to predict actions directly from egocentric observations, inspiring subsequent extensions~\cite{cheng2024navila,wei2025streamvln,zhang2024uninavid}.
However, the absence of environment priors often limits robustness in complex scenarios~\cite{hong2025general}.
% other methods
To address this, other methods incorporate explicit environmental representations, including topological maps~\cite{an2024etpnav,li2025ground,an2022bevbert,chen2022think,shah2023lm,chiang2024mobility,zhou2024navgpt}, value maps~\cite{long2024instructnav}, semantic maps~\cite{zhang2025mapnav}, and frontier maps~\cite{zhang2025mem2ego}.
While these works improve instruction-following, the reliance on rigid, step-by-step guidance differs from goal-oriented navigation settings that require flexible planning.
% goal-directed object navigation
Goal-directed object navigation~\cite{qi2020reverie,zhu2021soon,hirose2024lelan,yokoyama2024hm3d} addresses this limitation by providing more natural, high-level goals rather than exhaustive stepwise instructions.
Emerging task settings further push VLN toward persistent and interactive applications, including long-horizon navigation~\cite{song2025towards,anwar2025remembr}, scene adaptation~\cite{hong2025general}, and multi-turn dialog navigation with clarifications~\cite{han2025dialnav,ramrakhya2025grounding}.
These advances highlight the importance of maintaining persistent spatial grounding and reasoning over extended sequences.
% Beyond instruction-following VLN, goal-oriented object navigation~\cite{qi2020reverie,zhu2021soon,hirose2024lelan,yokoyama2024hm3d} and emerging settings such as scene adaptation~\cite{hong2025general}, dialog-based navigation~\cite{han2025dialnav,ramrakhya2025grounding}, and long-horizon VLN~\cite{song2025towards,anwar2025remembr,song2025longhorizon,zuo2026fantasyvln,wang2026vlingnav} further expand the task scope toward realistic decision-making.
The work most closely related to VideoNIG is Mobility VLA~\cite{chiang2024mobility}, which also leverages tour videos and user queries.
However, it relies on discrete topological graphs for localization and planning.
In contrast, VideoNIG treats continuous ego-centric tour videos as dense video-native priors, enabling flexible goal-oriented instruction generation without explicit graph abstraction.

\noindent \textbf{Navigation Instruction Generation} (NIG) has long been regarded as an auxiliary task for VLN, \ie , trajectory captioning.
% speaker
Speaker models~\cite{fried2018speaker,tan2019learning,wang2023scaling} generate instructions from route observations to augment navigation training data.
YTB-VLN~\cite{lin2023learning} and RoomTour3D~\cite{han2025roomtour3d} leverage online room tour videos to generate human trajectories and corresponding instructions.
LANA~\cite{wang2023lana} and CCC-VLN~\cite{wang2022counterfactual} learn instruction following (navigation) and generation in a single model.
ASSISTER~\cite{huang2022assister} generates instructions conditioned on planned routes for blind followers.
VLN-SRDF~\cite{wang2024bootstrapping} iteratively refines the instruction generator and navigator.
NaviLLM~\cite{zheng2024navillm} incorporates the trajectory summarization and EQA~\cite{wijmans2019embodied} into multi-task learning.
FCA-NIG~\cite{cui2025generating} considers sub-instruction-level trajectory alignment.
Training-free pathway instruction generation method~\cite{dorbala2024can} employs VQA and LLMs by in-context learning.
To better incorporate spatial cues, explicit environment representations are commonly used, including landmarks~\cite{wang2022less}, panoramic views~\cite{wang2022counterfactual}, hierarchical trees~\cite{wang2025navrag}, BEV features~\cite{fan2024navigation}, and map context~\cite{fan2025scene}.
% decider
Recently, NIG is adopted for the navigation task directly, such as intermediate state reasoning~\cite{zhou2024navgpt} and goal-conditioned visual navigation~\cite{wu2025govig}.
VideoNIG differs from these works by using tour videos as compact environmental priors for goal-oriented NIG.

\noindent \textbf{Visual Spatial Intelligence} is emerging as a key research focus within the multimodal large language model community, laying the groundwork for embodied intelligence in the physical world.
VSI-Bench~\cite{yang2025thinking} presents a video-based benchmark with various space-aware tasks, including route planning that highlights direction choices.
VLM4D~\cite{zhou2025vlm4d} benchmarks the spatio-temporal reasoning capabilities of vision-language models.
LLaVA-AURORA~\cite{bigverdi2025perception} and HiMTok~\cite{wang2025himtok} demonstrate the potential of MLLMs to exhibit strong foundational abilities in depth estimation, referring expression segmentation and object localization.
VGGT~\cite{wang2025vggt} infers 3D attributes of scenes from single or multiple RGB views.
3DRS~\cite{huang2025mllms} improves 3D representations of MLLMs by distilling from 3D foundation models.
Recent open-source MLLMs, such as InternVL3.5~\cite{wang2025internvl35} and Qwen3-VL~\cite{bai2025qwen3vl}, achieve notable progress in visual spatial understanding.
These advancements have made the task of generating navigation instructions directly from videos a promising and compelling research direction.

\begin{figure*}[t]
    \centering
    \includegraphics[width=1.0\linewidth]{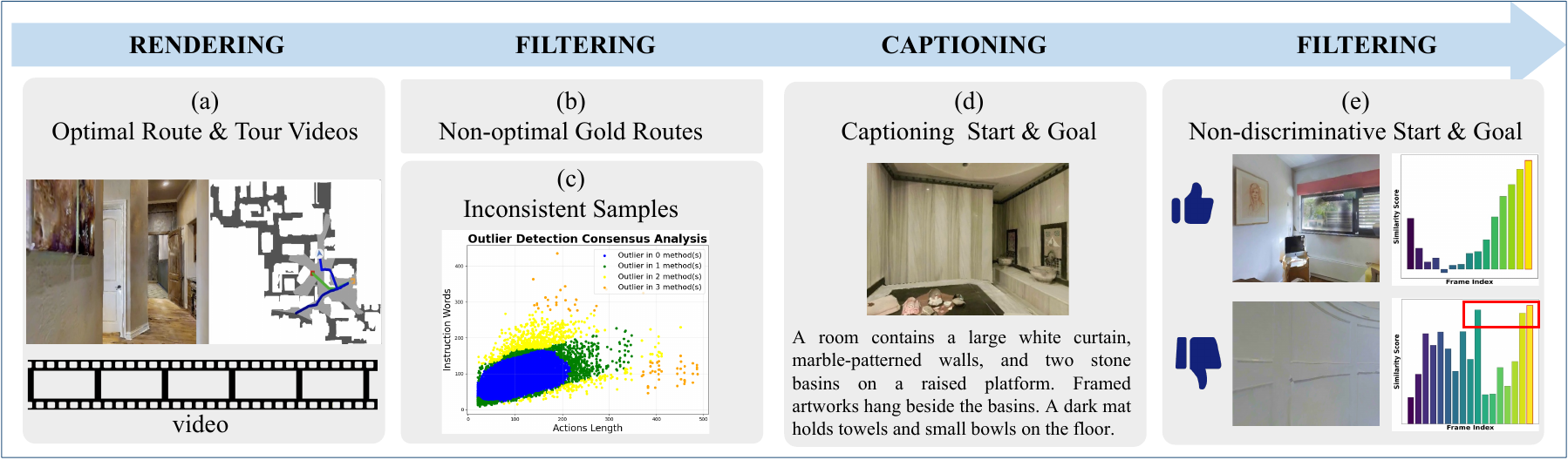}
    \caption{
        Benchmark construction pipeline for VideoNIG.
        (a) The left panel shows the observation at the arrow position in the right panel. The orange and red points are the initial and goal positions of the green Gold Route, respectively, and the blue trajectory is an Explorer Tour.
        (c) Outliers are filtered via three different methods. Blue and green points remain after filtering.
        (e) The initial and goal descriptions (text or image) are supposed to be discriminative with other scene observations.
    }
    \vspace{-3mm}
    \label{fig:data_pipe}
\end{figure*}

\FloatBarrier
\section{VideoNIG Task and Benchmark Instantiation}
\label{sec:dataset}
\begin{figure*}[t]
    \centering
    \includegraphics[width=1.0\linewidth]{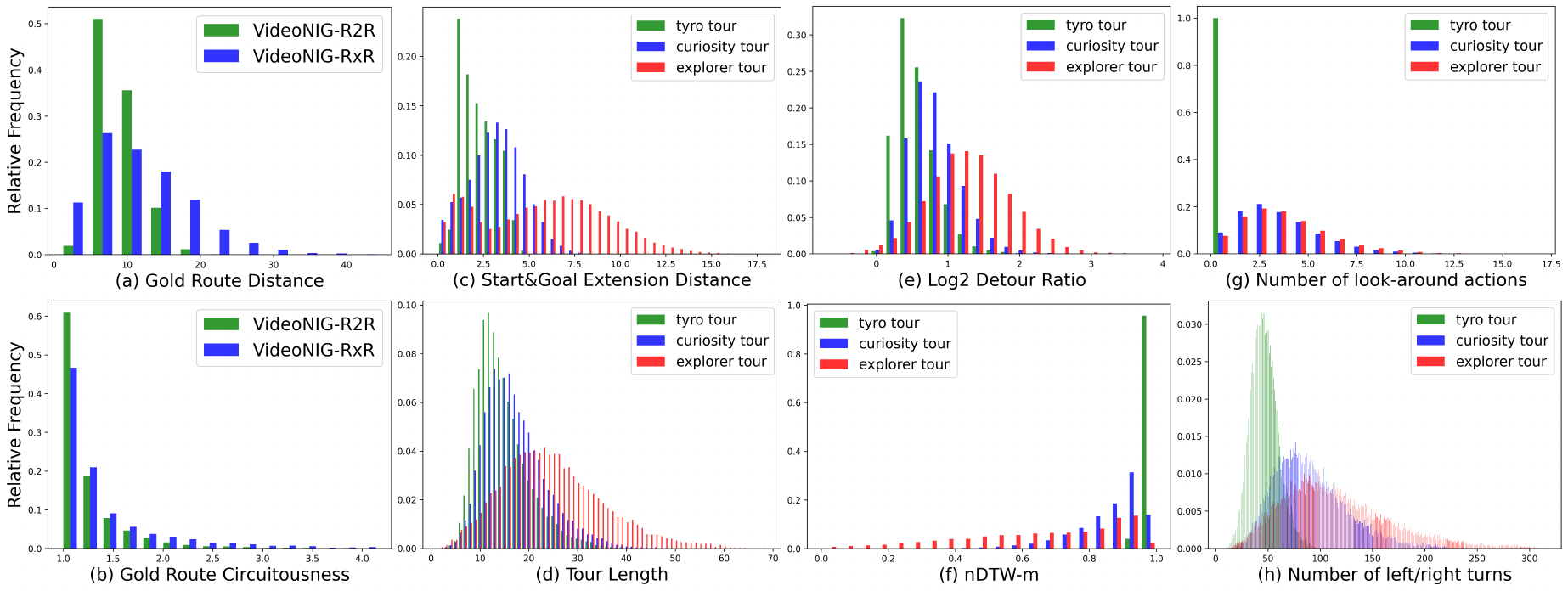}
    \vspace{-2mm}
    \caption{
        Data distributions in VideoNIG.
        (a) and (b) compare the distance and circuitousness of Gold Route across the filtered R2R and RxR samples.
        (c)-(h) compare the distributional differences among different tours.
    }
    \vspace{-4mm}
    \label{fig:data_hist}
\end{figure*}

\subsection{Task Definition}
The proposed VideoNIG task aims to generate fine-grained navigation instructions $\mathcal{I}$ for optimal routes conditioned on goal descriptions $\mathcal{G}$, initial RGB observations $\mathcal{O}_0$, and relevant tour videos $\mathcal{V}$, \ie , $\mathbf{P}\left(\mathcal{I}|\mathcal{G}, \mathcal{O}_0, \mathcal{V}\right)$.
The tour video captures environmental priors along the optimal route, including the start and end waypoints, while potentially incorporating additional observations beyond this route.
Goal descriptions may consist of text and images.
% The goal descriptions can be pure text, goal images, or a combination of both.
% This process can be formulated as:
% \begin{equation}
%     \begin{aligned}
%         \mathbf{P}\left(\mathcal{I}|\mathcal{G}, \mathcal{O}_0, \mathcal{V}\right)
%     \end{aligned}
% \end{equation}

% TODO: post the prompt?

\subsection{Benchmark Instantiation}
We instantiate VideoNIG based on the classical R2R-CE and RxR-CE~\cite{anderson2018r2r,ku2020rxr,krantz2020vlnce} in continuous environments through the Habitat simulator~\cite{savva2019habitat}, using their high-quality human-annotated navigation instructions.
This instantiation serves as a controlled diagnostic setting for studying the VideoNIG task; it is not intended to claim coverage of naturally collected tour videos or human-aware/social navigation scenarios.
We construct the benchmark through a carefully designed pipeline to ensure the quality of the tour videos and navigation instructions, as well as the distinctiveness of initial and goal descriptions, as shown in \cref{fig:data_pipe}.
See \cref{sec: tour_details} for more details.

\noindent\textbf{Rendering Gold Route \& Tour Videos.}
We render four types of videos in the simulator.
First, the original annotated route waypoints are used for rendering, yielding \textit{Gold Route} videos.
These videos correspond directly to their annotated instructions.
We then prepare \textit{Tyro Tour} videos by extending the Gold Route with a few random steps at both the start and goal points.
In addition, \textit{Curiosity Tour} videos are rendered by introducing two random behaviors: looking around and visiting neighboring areas.
Furthermore, \textit{Explorer Tour} videos are generated by applying larger offsets to intermediate points on the Gold Route, resulting in lower trajectory overlap.
Each frame is captured after a discrete action is taken.
The video fps is 6.
Strict connectivity checks, obstacle avoidance, and manual inspection ensure video quality.

\noindent \textbf{Filtering Non-optimal Routes.}
Some annotated routes in the original R2R-CE and RxR-CE datasets are not optimal from the start to the goal, which is not expected in VideoNIG.
We additionally calculate the shortest trajectories in the simulator and filter out suboptimal Gold Routes using nDTW~\cite{ilharco2019ndtw}, a metric for measuring trajectory similarity.
Routes with nDTW $< 0.8$ are regarded as non-optimal and removed.

\noindent \textbf{Filtering Inconsistent Samples.}
Based on our observations, some samples contain incomplete, coarse, or overly redundant instructions, as well as abnormal trajectories caused by flaws in the original annotations.
Approximately one-third of the original instructions omit a necessary ``turn around'' action at the beginning, compared with the action sequences.
% We complete these instructions using Qwen3-30B-A3B~\cite{qwen3}.
To ensure alignment between trajectory complexity and linguistic descriptions, as well as temporal consistency in tours, we employ an ensemble outlier-detection strategy that integrates multiple methods to assess two critical correlations: action sequence length $\left|\mathcal{A}\right|$ versus instruction word count $\left|\mathcal{I}\right|$, and $\left|\mathcal{A}\right|$ versus tour video duration $\left|\mathcal{V}\right|$.
We use three methods for ensembling -- Z-score, RANSAC~\cite{fischler1981random}, and Isolation Forest~\cite{liu2008isolation} -- and retain samples that are identified as outliers by at most one method across both correlations.
\cref{fig:data_pipe}(c) shows the $\left|\mathcal{I}\right|$-$\left|\mathcal{A}\right|$ scatter plot, where the blue and green points are retained while the others are dropped.

\noindent \textbf{Captioning Initial \& Goal Observations.}
To obtain multimodal descriptions for the start and goal points, we caption the corresponding RGB views using the \texttt{qwen3-vl-plus} API.
The descriptions are required to be concise and clear.

\noindent \textbf{Filtering Non-discriminative Initial \& Goal views.}
Non-discriminative initial and goal descriptions can easily confuse the agent during localization.
We filter~\cite{liu2025lamra} them using text-image and image-image similarities between the start or goal descriptions and other observation frames in Gold Route videos (\cref{fig:data_pipe}(e)).

% \noindent \textbf{Filling Multimodal Prompt.}

\subsection{Statistics}
% The VideoNIG dataset consists of around 20K gold route videos derived from R2R-CE and RxR-CE, and 60K tour videos, as well as 37K unambiguous multimodal start and goal descriptions.
% Details are listed in \cref{tab:dataset_num}.
% The training and val\_unseen sets are split following the configuration of the original datasets.
The VideoNIG dataset contains approximately 20K Gold Route videos derived from R2R-CE and RxR-CE, 60K tour videos, and 37K unambiguous multimodal start–goal descriptions (\cref{tab:dataset_num}). The training and val\_unseen splits follow the original dataset configurations.
\begin{table}[t]
    \centering
    \caption{Statistics of videos and goals in VideoNIG. Each cell lists the numbers in the training and val\_unseen sets.}
    \vspace{-2mm}
    \scriptsize
    \renewcommand{\arraystretch}{0.9}
    \begin{tabular*}{\linewidth}{@{\extracolsep{\fill}}l|cc|cc|cc@{}}
        \toprule[0.8pt]
        Data source & \multicolumn{2}{c}{R2R-CE} & \multicolumn{2}{c}{RxR-CE} & \multicolumn{2}{c}{Total} \\
        \cmidrule(lr){2-3} \cmidrule(lr){4-5} \cmidrule(lr){6-7}
        & Train & Val\_unseen & Train & Val\_unseen & Train & Val\_unseen \\
        \midrule[0.5pt]
        Gold Route  & 7778 & 1341 & 9623 & 1962 & 17401 & 3303 \\
        Tyro Tour  & 7441 & 1272 & 9395 & 1840 & 16836 & 3112 \\
        Curiosity Tour  & 7418 & 1256 & 9381 & 1849 & 16799 & 3105 \\
        Explorer Tour & 7358 & 1233 & 9367 & 1792 & 16725 & 3025 \\
        \midrule[0.5pt]
        Image Goal & 7586 & 1313 & 9408 & 1934 & 16994 & 3247 \\
        Textual Goal & 6407 & 1183 & 7628 & 1629 & 14035 & 2812 \\
        \bottomrule[0.8pt]
    \end{tabular*}
    \vspace{-4mm}
    \label{tab:dataset_num}
\end{table}

% \begin{table}[t]
%     \centering
%     \caption{Statistics about videos and goals in VideoNIG. Each cell lists the numbers in training and val\_unseen set.}
%     \label{tab:dataset_num}
%     % \scriptsize
%     % \small
%     \fontsize{6.5pt}{6.5pt}\selectfont
%     % \renewcommand{\arraystretch}{0.9}
%     \begin{tabular}{@{}lcc|cc|cc@{}}
%         \toprule[0.8pt]
%         Data source & \multicolumn{2}{c}{R2R-CE} & \multicolumn{2}{c}{RxR-CE} & \multicolumn{2}{c}{Total} \\
%         \cmidrule(lr){2-3} \cmidrule(lr){4-5} \cmidrule(lr){6-7}
%         & Train & Val\_unseen & Train & Val\_unseen & Train & Val\_unseen \\
%         \midrule[0.5pt]
%         Gold Route      & 7778 & 1341 & 9623 & 1962 & 17401 & 3303 \\
%         Tyro Tour       & 7441 & 1272 & 9395 & 1840 & 16836 & 3112 \\
%         Curiosity Tour  & 7418 & 1256 & 9381 & 1849 & 16799 & 3105 \\
%         Explorer Tour   & 7358 & 1233 & 9367 & 1792 & 16725 & 3025 \\
%         \midrule[0.5pt]
%         Image Goal      & 7586 & 1313 & 9408 & 1934 & 16994 & 3247 \\
%         Textual Goal    & 6407 & 1183 & 7628 & 1629 & 14035 & 2812 \\
%         \bottomrule[0.8pt]
%     \end{tabular}
% \end{table}

% difficulty levels of gold route videos.
\noindent \textbf{Difficulty Levels of Gold Route.}
% The filtered gold routes from R2R and RxR preserve distribution patterns similar to their original datasets~\cite{ku2020rxr}.
% The average \textit{route distances} in VideoNIG\_RxR are longer than VideoNIG\_R2R (see \cref{fig:data_hist}(a)).
% To quantify the tortuosity of the gold routes, we compute \textit{circuitousness} as the ratio between the route distance and the straight-line Euclidean distance between the start and goal.
% Small values closer to 1 indicate a nearly straight route, generally corresponding to low difficulty.
% As shown in \cref{fig:data_hist}(b), gold routes in VideoNIG\_RxR are on average more tortuous, resulting in a more challenging task.
The filtered routes preserve the original distribution characteristics~\cite{ku2020rxr}.
VideoNIG-RxR exhibits longer average route distances than VideoNIG-R2R (\cref{fig:data_hist}(a)).
To measure route complexity, we define \textit{circuitousness} as the ratio between the route distance and the straight-line Euclidean distance between start and goal. Values closer to 1 indicate straighter and easier routes. As shown in \cref{fig:data_hist}(b), VideoNIG-RxR routes are generally more tortuous, leading to higher difficulty.

% difficulty levels of tour videos.
\noindent \textbf{Distribution of tour videos.}
We analyze tour-video distributions from six perspectives.
(1) Extended distances from the Gold Route endpoints.
Longer extensions provide broader and more redundant environmental scenes for optimal route planning.
Note that the peak near distance=0 for Explorer Tour in \cref{fig:data_hist}(c) occurs because some long-extended points are unreachable, prompting us to narrow the extension range.
(2) Tour lengths.
(3) Logarithmic detour ratios, calculated as the log ratio between the tour length and the optimal Gold Route length.
% M-nDTW -> nDTW_m
(4) nDTW-m, the nDTW metric between the Gold Route and the matched sub-trajectory from the tour that aligns the start and goal points.
(5) Number of look-around actions.
Looking around during tours captures more environmental information, but also introduces more distractions for temporal video understanding.
(6) Number of left and right turns.
% As illustrated in \cref{fig:data_hist}(c)-(h), Tyro Tour that simply extend the endpoints in the Gold Route exhibits the most concentrated distribution.
As illustrated in \cref{fig:data_hist}(c)-(h), Tyro Tour exhibits the most concentrated distributions.
% The distribution for Explorer Tour shows the longest extensions and trajectories, the most route deviation, the least overlap with Gold Route, and the highest number of left\&right turning actions.
Conversely, Explorer Tour represents the most challenging scenarios, characterized by the greatest trajectory lengths, route deviations, and turning frequencies.
The look-around action distributions for Curiosity Tour and Explorer Tour are similar.

% \subsection{Task Evaluation Metric}
\subsection{Task Evaluation}
% summary of evaluation metrics
To comprehensively evaluate the quality and utility of generated navigation instructions, we adopt a three-pronged evaluation protocol.

\noindent \textbf{Text Similarity Evaluation. }Assessing the effectiveness of navigation instructions is essential and non-trivial.
% The main challenge lies in the inherent diversity of navigation route descriptions, including variations in the selection of key waypoints, the choice of contextual scene elements, and linguistic expression.
Inspired by~\cite{zhao2021evaluation}, we perform system-level evaluation using Rouge-L~\cite{lin2004rouge} and SPICE~\cite{anderson2016spice}, which measure text similarity between the annotated and generated instructions.
\begin{figure}[t]
    \centering
    \includegraphics[width=1\linewidth]{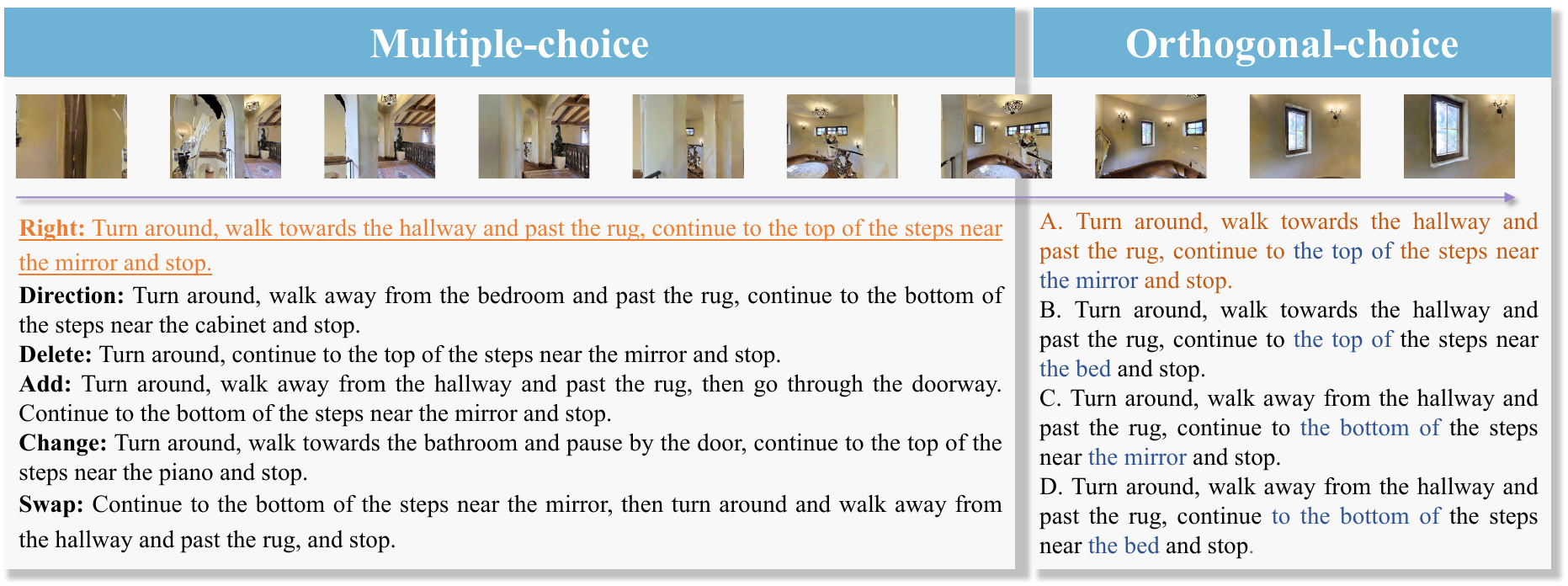}
    \vspace{-2mm}
    \caption{
        % Multiple-choice and Orthogonal-choice.
        Multiple-choice: tests spatio-temporal consistency via diverse distractors. Orthogonal-choice: isolates direction and entity changes to mitigate textual shortcuts.
    }
    \vspace{-4mm}
    \label{fig:neg_choice}
\end{figure}
% 这里加上负样本相关的制作和评测
% To evaluate whether the model captures spatial relationships in videos, we construct a choice-based benchmark inspired by NavBench~\cite{qiao2025navbench}. Six types of negative samples are designed: changing direction, changing entity words, adding or deleting subpaths, swapping subpath order, and altering subpaths.
% Since MPO training may lead the model to exploit superficial textual patterns, we further introduce an orthogonal benchmark that modifies only directions and entities (\cref{fig:neg_benchmark}), reducing shortcut exploitation. All benchmarks are evaluated using Accuracy.

\noindent \textbf{Choice Evaluation.} To rigorously assess whether the model accurately captures spatial relationships in tour videos, we design a Choice Evaluation inspired by NavBench~\cite{qiao2025navbench}. It contains two parts: Multiple-choice and Orthogonal-choice.
Specifically, Multiple-choice employs five types of negative distractors (\cref{fig:neg_choice}), including direction and entity modifications, and various subpath modifications (deletion, addition, change, and swapping).
These distractors explicitly examine fine-grained spatio-temporal consistency between generated instructions and the corresponding video trajectories.
Complementary to this, Orthogonal-choice focuses exclusively on direction and entity modifications (\cref{fig:neg_choice}) to mitigate reliance on superficial textual patterns.
Performance is measured by the accuracy of selecting the ground-truth description among the choices.

% dualvln evaluation.
% \noindent In addition to the above evaluations, we further feed the generated instructions into an off-the-shelf VLN agent to see the effect for real navigation performance based on ~\cite{wei2025ground}, despite incurring higher evaluation overhead.
\noindent \textbf{Navigation Execution. }Beyond text-based and discriminative evaluations, we further assess downstream navigation performance by feeding the generated instructions into an off-the-shelf VLN agent~\cite{wei2025ground,internvla-n1}.
This setup evaluates navigation execution performance while explicitly decoupling instruction quality from execution bias, providing a complementary perspective despite the higher computational overhead.
% Beyond the single-modality evaluation, we further feed the generated instructions into an off-the-shelf VLN agent to see the effect for real navigation performance, despite incurring higher evaluation overhead.
% add negative samples evaluation and navigation execution results
% Inspired by ~\cite{zhao2021evaluation}, we evaluate from two aspects. One is for the \textit{text similarity} between the annotated and generated instructions, including SPICE~\cite{anderson2016spice} and Rouge-L~\cite{lin2004rouge}.
% Another is \textit{vl-score} judged by a novel vision-language model-based approach, VLScorer, that is built upon the MLLM Qwen3-VL-8B~\cite{bai2025qwen3vl}, which we select based on its strong empirical performance in the tour video caption task.
% Given a gold route video and a navigation instruction, VLScorer is trained to evaluate the instruction along two dimensions: path consistency, which measures the alignment between the instruction and the gold route video, and expressiveness, which reflects the clarity and conciseness of the textual description. Both dimensions are represented as continuous scores within the range [0, 1].
% The scoring prompt is listed in \cref{supp:vlscorer}.
% The final score is the average of SPICE, Rouge-L, path consistency and expressiveness.

% TODO: explain why MLLM works for scoring

\FloatBarrier
% \section{VideoNIG-Method}
% 基于课程学习的方法
\section{Method}
\label{sec:method}
We adopt a two-stage Curriculum Learning (CL)~\cite{bengio2009curriculum} paradigm to stabilize optimization and progressively enhance the model's spatial-semantic reasoning capability.
Unlike conventional training that randomly mixes samples of varying complexity, our curriculum explicitly decomposes the learning process into two dimensions: Action Warmup and Complexity Progression. Formally, given a dataset $D=\{(x_i,y_i)\}_{i=1}^N$ and loss function $\mathcal{L}$, Curriculum Learning can be viewed as a stage-wise optimization process: $L(t)=\sum_{i=1}^{N} w_i^{(t)} \cdot \mathcal{L}(f(x_i;\theta), y_i)$, where $w_i^{(t)}$ controls the contribution of sample $i$ at training stage $t$.
In our setting, the curriculum is instantiated through structured stage-wise data scheduling rather than explicit re-weighting.

\noindent \textbf{Stage 1: Action Warmup.}
We prime the spatial reasoning capability of Qwen3-VL-8B-Instruct through an action-view alignment stage.
This stage includes two objectives: action-from-view (predicting relative motion between two frames) and view-from-action (selecting the resulting frame given a specified action).
By focusing on local spatial transformations, this stage strengthens orientation sensitivity before introducing long-range temporal dependencies.
Details of data construction and training for Action Warmup are provided in \cref{sec: action_warmup}.
% We prime the spatial reasoning capacity of Qwen3-VL through an action–view alignment stage. 
% This stage comprises two discriminative objectives: (i) \textit{action-from-view}, where the model identifies the relative motion (e.g., forward $x$ meters, turn left/right $\theta$ degrees) between two frames; and (ii) \textit{view-from-action}, which requires selecting the correct destination frame from four candidates given a source frame and an action command. 
% By focusing on these local transformations, this stage instills robust orientation sensitivity before the model is tasked with capturing complex, long-range temporal dependencies in full video trajectories.

\noindent \textbf{Stage 2: Complexity Progression.}
We adopt a Curriculum Learning strategy that transitions from low-complexity trajectories (Gold Route and Tyro Tour) to higher-complexity ones (Curiosity Tour and Explorer Tour).
Complex data are introduced after stable convergence on simpler navigation semantics, promoting smoother optimization and improving long-horizon reasoning robustness.

\FloatBarrier
\section{Experiments}
\label{sec:exp}
Our experiments are organized around four questions: task difficulty, diagnostic validity, curriculum effectiveness, and downstream executability.
% \subsection{Training}
\subsection{Training and Implementation Details.}
\noindent \textbf{Model Training.}
We use Qwen3-VL-8B-Instruct as the base model.
We implement two training methods: Group Relative Policy Optimization (GRPO)~\cite{shao2024deepseekmath} and Mixed Preference Optimization (MPO)~\cite{wang2024enhancing}. Both are trained with the proposed Curriculum Learning schedule.
% 简单提一下奖励设置，主要放到附录
For GRPO, the reward function is a weighted combination of four complementary components:
(1) Instruction Similarity, measured by SPICE and Rouge-L scores.
(2) Linguistic Quality, evaluated by an LLM judge for clarity and relevance.
(3) Trajectory Consistency, quantified by an MLLM conditioned on the predicted instruction and Gold Route.
(4) Length Penalty to discourage redundancy.
MPO employs preference-based learning using negative samples constructed via our evaluation protocol (see \cref{fig:neg_choice}), where one negative sample is randomly selected per instance.
Details of the GRPO reward function design and MPO dataset construction are provided in \cref{sec: training_settings}.

\noindent \textbf{Training Details.}
% Stage 1: action warmup
The baseline VideoNIG model is trained by supervised fine-tuning (SFT) on Qwen3-VL-8B-Instruct.
Following~\cite{wang2025himtok}, we adopt a mixed data recipe for SFT to avoid overfitting, including all VideoNIG training samples, text datasets on math and code~\cite{conover2023free,zheng2024opencodeinterpreter,yue2023mammoth,yu2023metamath,chen2024allava,mitra2024orca}, and several general multimodal datasets~\cite{han2025roomtour3d,ouyang2025spacer,chen2024allava,wang2024cogvlm}.
The ratio between VideoNIG data and other data is about 1:1.
During SFT, only the vision module is frozen, while all other components are fine-tuned for 1 epoch.
The frame sampling rate is set to 6, with a maximum of 210 frames.
The learning rate is initialized to 1e-5 and uses a cosine scheduler.
% The SFT for VLScorer is conducted on 8 Nvidia A100 GPUs for 5 hours. 
% Action Warmup 
The Action Warmup stage is optimized using GRPO to strengthen orientation discrimination before full instruction learning.
Training is conducted on 4 NVIDIA A100 GPUs for 24 hours with a constant learning rate of $1 \times 10^{-6}$.
%GRPO
During GRPO, the learning rate is kept constant at 2e-6.
The KL coefficient is set to 5e-3, and $\varepsilon_{\text{high}}$ is set to 0.28 to encourage exploration.
GRPO training is conducted on 8 NVIDIA A100 GPUs for 12 hours.
% MPO
During MPO, the learning rate is kept constant at 5e-6. Training is conducted on 4 NVIDIA A100 GPUs for 12 hours.
% MPO lora
After MPO, we further perform simple LoRA fine-tuning with a learning rate of 1e-4 for 1 epoch.

% The baseline VideoNIG model is also trained by SFT on Qwen3-VL.
% All the VideoNIG training samples are used, along with other multimodal and text datasets as mixed in tuning VLScorer.

\begin{table*}[t]
    \centering
    \caption{Evaluation results on VideoNIG.}
    \vspace{-2mm}
    \label{tab:all_results}
    \resizebox{\linewidth}{!}{
        \begin{tabular}{l|cc|cc|cc|cc|cc|cc|cc|cc|c}
            \toprule[1.1pt]
            Video type & \multicolumn{4}{c|}{Gold Route} & \multicolumn{4}{c|}{Tyro Tour} & \multicolumn{4}{c|}{Curiosity Tour} & \multicolumn{4}{c|}{Explorer Tour} & \multirow{3}{*}{Average} \\
            \cline{1-17}
            Source & \multicolumn{2}{c|}{R2R} & \multicolumn{2}{c|}{RxR} & \multicolumn{2}{c|}{R2R} & \multicolumn{2}{c|}{RxR} & \multicolumn{2}{c|}{R2R} & \multicolumn{2}{c|}{RxR} & \multicolumn{2}{c|}{R2R} & \multicolumn{2}{c|}{RxR} & \\
            \cline{1-17}
            Goal modal & image & text & image & text & image & text & image & text & image & text & image & text & image & text & image & text & \\
            \midrule[0.7pt]
            \multicolumn{18}{l}{Text Similarity: Rouge-L on top, SPICE below.} \\
            \midrule[0.7pt]
            % InternVL3-8B~\cite{zhu2025internvl3} & \makecell{0.160\\0.087} & \makecell{0.165\\0.087} & \makecell{0.173\\0.117} & \makecell{0.173\\0.112} & \makecell{0.158\\0.088} & \makecell{0.163\\0.086} & \makecell{0.174\\0.114} & \makecell{0.174\\0.113} & \makecell{0.157\\0.086} & \makecell{0.163\\0.086} & \makecell{0.173\\0.116} & \makecell{0.173\\0.109} & \makecell{0.153\\0.085} & \makecell{0.162\\0.085} & \makecell{0.172\\0.114} & \makecell{0.174\\0.111} & \makecell{0.167\\0.100} \\
            % \hline
            % InternVL3.5-4B~\cite{wang2025internvl35} & \makecell{0.191\\0.105} & \makecell{0.174\\0.097} & \makecell{0.173\\0.116} & \makecell{0.174\\0.115} & \makecell{0.189\\0.101} & \makecell{0.170\\0.090} & \makecell{0.174\\0.116} & \makecell{0.173\\0.114} & \makecell{0.188\\0.101} & \makecell{0.169\\0.091} & \makecell{0.174\\0.116} & \makecell{0.172\\0.112} & \makecell{0.187\\0.101} & \makecell{0.168\\0.089} & \makecell{0.172\\0.116} & \makecell{0.173\\0.111} & \makecell{0.176\\0.106} \\
            % \hline
            InternVL3.5-8B~\cite{wang2025internvl35} & \makecell{0.217\\0.109} & \makecell{0.212\\0.105} & \makecell{0.174\\0.112} & \makecell{0.176\\0.114} & \makecell{0.213\\0.109} & \makecell{0.210\\0.102} & \makecell{0.174\\0.113} & \makecell{0.177\\0.114} & \makecell{0.212\\0.106} & \makecell{0.209\\0.103} & \makecell{0.174\\0.113} & \makecell{0.176\\0.112} & \makecell{0.209\\0.108} & \makecell{0.206\\0.100} & \makecell{0.174\\0.111} & \makecell{0.176\\0.113} & \makecell{0.193\\0.109} \\
            \hline
            % Qwen2.5-VL-7B~\cite{bai2025qwen2_5} & \makecell{0.195\\0.109} & \makecell{0.182\\0.094} & \makecell{0.164\\0.101} & \makecell{0.171\\0.102} & \makecell{0.199\\0.106} & \makecell{0.178\\0.095} & \makecell{0.166\\0.097} & \makecell{0.175\\0.104} & \makecell{0.198\\0.106} & \makecell{0.178\\0.093} & \makecell{0.166\\0.099} & \makecell{0.173\\0.102} & \makecell{0.195\\0.105} & \makecell{0.177\\0.094} & \makecell{0.166\\0.100} & \makecell{0.175\\0.105} & \makecell{0.179\\0.101} \\
            % \hline
            % Qwen3-VL-4B~\cite{bai2025qwen3vl} & \makecell{0.224\\0.114} & \makecell{0.201\\0.097} & \makecell{0.172\\0.131} & \makecell{0.181\\0.118} & \makecell{0.169\\0.065} & \makecell{0.097\\0.053} & \makecell{0.187\\0.095} & \makecell{0.195\\0.136} & \makecell{0.243\\0.107} & \makecell{0.201\\0.091} & \makecell{0.192\\0.132} & \makecell{0.180\\0.116} & \makecell{0.217\\0.110} & \makecell{0.196\\0.091} & \makecell{0.201\\0.141} & \makecell{0.178\\0.113} & \makecell{0.190\\0.107} \\
            Qwen3-VL-8B~\cite{bai2025qwen3vl} & \makecell{0.213\\0.121} & \makecell{0.225\\0.098} & \makecell{0.175\\0.111} & \makecell{0.176\\0.107} & \makecell{0.212\\0.117} & \makecell{0.201\\0.101} & \makecell{0.176\\0.111} & \makecell{0.175\\0.104} & \makecell{0.210\\0.117} & \makecell{0.199\\0.100} & \makecell{0.175\\0.109} & \makecell{0.174\\0.103} & \makecell{0.212\\0.116} & \makecell{0.199\\0.096} & \makecell{0.173\\0.107} & \makecell{0.174\\0.102} & \makecell{0.192\\0.107} \\
            % \midrule[0.7pt]
            % Qwen3-VL-4B (SFT) & \makecell{0.277\\0.156} & \makecell{0.282\\0.154} & \makecell{0.201\\0.157} & \makecell{0.202\\0.156} & \makecell{0.267\\0.148} & \makecell{0.269\\0.151} & \makecell{0.202\\0.158} & \makecell{0.201\\0.155} & \makecell{0.254\\0.147} & \makecell{0.257\\0.143} & \makecell{0.200\\0.155} & \makecell{0.200\\0.155} & \makecell{0.240\\0.134} & \makecell{0.243\\0.138} & \makecell{0.199\\0.154} & \makecell{0.197\\0.151} & \makecell{0.231\\0.151} \\
            \hline
            Qwen3-VL-8B (SFT) & \makecell{0.283\\0.156} & \makecell{0.284\\0.159} & \makecell{0.197\\0.151} & \makecell{0.197\\0.152} & \makecell{0.274\\0.152} & \makecell{0.278\\0.156} & \makecell{0.195\\0.147} & \makecell{0.196\\0.148} & \makecell{0.267\\0.149} & \makecell{0.271\\0.146} & \makecell{0.195\\0.148} & \makecell{0.193\\0.147} & \makecell{0.258\\0.143} & \makecell{0.259\\0.139} & \makecell{0.193\\0.146} & \makecell{0.191\\0.143} & \makecell{0.233\\0.149} \\
            \hline
            Qwen3-VL-8B (GRPO) & \makecell{0.301\\0.196} & \makecell{0.299\\0.195} & \makecell{0.209\\0.194} & \makecell{0.209\\0.190} & \makecell{0.301\\0.193} & \makecell{0.299\\0.193} & \makecell{0.211\\0.195} & \makecell{0.210\\0.189} & \makecell{0.298\\0.194} & \makecell{0.297\\0.190} & \makecell{0.211\\0.196} & \makecell{0.210\\0.191} & \makecell{0.296\\0.190} & \makecell{0.294\\0.185} & \makecell{0.211\\0.189} & \makecell{0.209\\0.186} & \makecell{0.254\\\textbf{0.192}} \\

            \hline
            Qwen3-VL-8B (MPO) & \makecell{0.348\\0.196} & \makecell{0.340\\0.187} & \makecell{0.186\\0.126} & \makecell{0.182\\0.121} & \makecell{0.334\\0.186} & \makecell{0.315\\0.170} & \makecell{0.191\\0.131} & \makecell{0.187\\0.133} & \makecell{0.296\\0.173} & \makecell{0.270\\0.151} & \makecell{0.179\\0.133} & \makecell{0.171\\0.125} & \makecell{0.266\\0.161} & \makecell{0.226\\0.136} & \makecell{0.166\\0.126} & \makecell{0.156\\0.123} & \makecell{0.238\\0.149} \\
            \hline
            Qwen3-VL-8B (MPO-lora) & \makecell{0.331\\0.190} & \makecell{0.335\\0.192} & \makecell{0.192\\0.149} & \makecell{0.182\\0.129} & \makecell{0.336\\0.195} & \makecell{0.338\\0.191} & \makecell{0.190\\0.144} & \makecell{0.185\\0.128} & \makecell{0.334\\0.189} & \makecell{0.327\\0.186} & \makecell{0.193\\0.147} & \makecell{0.183\\0.128} & \makecell{0.326\\0.185} & \makecell{0.306\\0.168} & \makecell{0.184\\0.136} & \makecell{0.182\\0.126} & \makecell{\textbf{0.258}\\0.161} \\
            \midrule[0.7pt]
            \multicolumn{18}{l}{Choice Evaluation: Multiple-choice on top, Orthogonal-choice below (accuracy \%).} \\
            \midrule[0.7pt]
            InternVL3.5-8B~\cite{wang2025internvl35} & \makecell{44.55\\49.54} & \makecell{49.40\\52.77} & \makecell{44.32\\50.78} & \makecell{44.89\\48.49} & \makecell{43.85\\48.35} & \makecell{45.19\\47.19} & \makecell{41.57\\45.42} & \makecell{44.76\\45.29} & \makecell{44.82\\49.51} & \makecell{45.03\\49.59} & \makecell{42.74\\45.78} & \makecell{44.24\\46.03} & \makecell{44.02\\48.55} & \makecell{46.33\\48.85} & \makecell{43.04\\45.36} & \makecell{42.88\\45.26} & \makecell{44.48\\47.92} \\
            \hline
            Qwen3-VL-8B~\cite{bai2025qwen3vl} & \makecell{54.95\\52.70} & \makecell{54.31\\55.24} & \makecell{47.45\\53.50} & \makecell{49.04\\54.42} & \makecell{52.86\\54.89} & \makecell{56.33\\51.34} & \makecell{47.85\\51.52} & \makecell{48.36\\50.92} & \makecell{55.44\\53.85} & \makecell{57.04\\52.81} & \makecell{46.13\\50.99} & \makecell{48.04\\49.41} & \makecell{52.50\\52.25} & \makecell{54.50\\52.04} & \makecell{46.66\\49.55} & \makecell{45.45\\48.72} & \makecell{51.06\\52.13} \\
            \hline
            Qwen3-VL-8B (SFT) & \makecell{58.00\\57.46} & \makecell{59.03\\58.28} & \makecell{44.42\\55.84} & \makecell{43.06\\53.42} & \makecell{60.82\\55.40} & \makecell{56.57\\58.18} & \makecell{42.78\\51.96} & \makecell{43.30\\50.92} & \makecell{56.98\\54.08} & \makecell{59.02\\55.25} & \makecell{41.50\\49.48} & \makecell{42.23\\47.55} & \makecell{53.49\\53.96} & \makecell{52.85\\50.83} & \makecell{42.16\\48.84} & \makecell{40.62\\48.31} & \makecell{49.80\\53.11} \\
            \hline
            Qwen3-VL-8B (GRPO) & \makecell{55.86\\59.50} & \makecell{56.29\\59.54} & \makecell{48.65\\54.78} & \makecell{49.35\\53.33} & \makecell{55.28\\56.57} & \makecell{54.69\\57.00} & \makecell{45.94\\52.07} & \makecell{47.73\\53.15} & \makecell{54.78\\57.10} & \makecell{56.94\\56.65} & \makecell{47.53\\52.20} & \makecell{46.58\\51.14} & \makecell{52.24\\56.81} & \makecell{53.54\\54.82} & \makecell{47.34\\48.61} & \makecell{44.65\\48.75} & \makecell{51.09\\54.50} \\
            \hline
            Qwen3-VL-8B (MPO) & \makecell{62.63\\68.05} & \makecell{65.54\\68.42} & \makecell{56.13\\59.13} & \makecell{56.42\\59.38} & \makecell{61.27\\66.59} & \makecell{61.07\\65.26} & \makecell{52.39\\59.25} & \makecell{51.99\\57.57} & \makecell{58.10\\59.07} & \makecell{55.23\\57.38} & \makecell{51.19\\56.15} & \makecell{49.61\\52.71} & \makecell{54.50\\56.69} & \makecell{49.62\\51.94} & \makecell{47.22\\50.86} & \makecell{47.04\\48.33} & \makecell{55.00\\58.55} \\
            \hline
            Qwen3-VL-8B (MPO-lora) & \makecell{65.56\\68.74} & \makecell{67.44\\70.66} & \makecell{62.14\\63.52} & \makecell{58.06\\62.38} & \makecell{64.80\\67.99} & \makecell{64.90\\67.21} & \makecell{59.11\\60.55} & \makecell{59.93\\58.78} & \makecell{63.51\\65.77} & \makecell{61.90\\62.81} & \makecell{58.91\\58.34} & \makecell{59.46\\57.26} & \makecell{59.47\\62.64} & \makecell{54.31\\57.43} & \makecell{55.98\\52.86} & \makecell{54.48\\54.55} & \makecell{\textbf{60.62}\\\textbf{61.97}} \\
            \bottomrule[1.1pt]
        \end{tabular}
    }
\end{table*}

\begin{figure*}[t]
    \centering
    \includegraphics[width=1\linewidth]{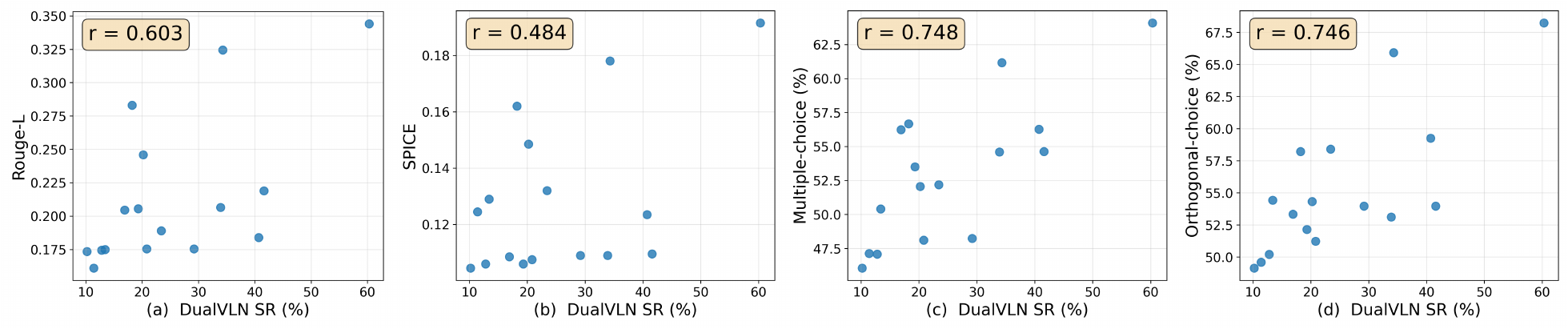}
    \vspace{-2mm}
    \caption{
        Correlation between Rouge-L, SPICE, Multiple-choice, Orthogonal-choice, and VLN Success Rate.
    }
    \vspace{-4mm}
    \label{fig:SR_correlation}
\end{figure*}

\noindent\textbf{Models and Evaluation.} 
% We select MLLMs that have demonstrated strong performance in visual spatial intelligence, including InternVL3~\cite{zhu2025internvl3}, InternVL3.5~\cite{wang2025internvl35}, Qwen2.5-VL~\cite{bai2025qwen2_5}, and Qwen3-VL~\cite{bai2025qwen3vl}, covering different scales and progressive versions.  
We select representative MLLMs with strong spatial performance, namely InternVL3~\cite{zhu2025internvl3}, InternVL3.5~\cite{wang2025internvl35}, Qwen2.5-VL~\cite{bai2025qwen2_5}, and Qwen3-VL~\cite{bai2025qwen3vl}.
All evaluations are conducted within the vLLM framework~\cite{kwon2023efficient}, using greedy decoding with temperature set to 0 and a video frame sampling rate of 4 fps, ensuring consistent and deterministic assessment across models.

\subsection{Results on VideoNIG}
% \subsubsection{Main Results}

\begin{figure*}[t]
    \centering
    \includegraphics[width=1\linewidth]{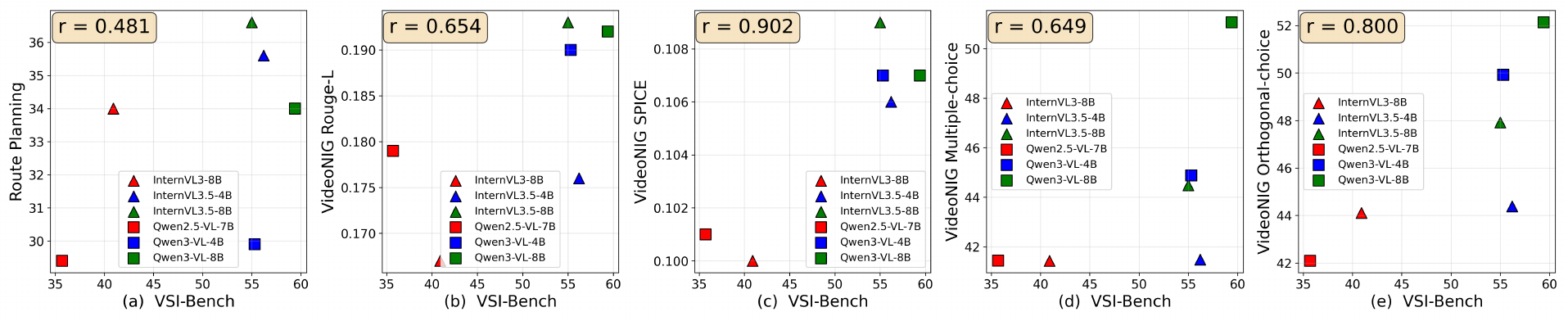}
    \vspace{-3mm}
    \caption{
        Correlation of VSI-Bench, Route Planning, and VideoNIG Performance
    }
    \vspace{-4mm}
    \label{fig:exp_vsi_nig}
\end{figure*}

\noindent \textbf{Main Results and Comparative Analysis. }
% As shown in \cref{tab:res_all_spice}, all models exhibit universally low Rouge-L and SPICE scores. 
% InternVL3.5-8B and Qwen3-VL-8B perform comparably, both improve over their earlier and 4B variants in Rouge-L, while their SPICE scores remain similar.
% Our supervised fine-tuned Qwen3-VL models show clear gains on both metrics, yet the performance gap between the 4B and 8B variants remains minor.
% Our CRL and MPO model achieves better performance, significantly surpassing the baseline Qwen3-VL-8B model. Current MLLMs also exhibit nearly no differences between the input image and textual goals, which supports the flexible goal description modals in real applications.
% As shown in , all models achieve relatively Rouge-L, SPICE, Multiple-choice and Orthogonal-choice, indicating the intrinsic difficulty of the VideoNIG task (Some models results are in Appendix). 
We evaluate InternVL and Qwen series models across four metrics: Rouge-L, SPICE, Multiple-choice, and Orthogonal-choice in \cref{tab:all_results}. Results for additional open-source and top-tier proprietary models are provided in \cref{sec:experimental_results}.

\noindent Regarding text similarity metrics, InternVL3.5-8B and Qwen3-VL-8B exhibit comparable performance, both outperforming their earlier or smaller variants in Rouge-L while maintaining similar SPICE scores.
Supervised fine-tuning of Qwen3-VL leads to consistent improvements across both metrics.
Specifically, both GRPO- and MPO-trained models significantly outperform the Qwen3-VL-8B baseline.
Notably, the GRPO-trained model achieves higher Rouge-L and SPICE scores, which can be attributed to the explicit alignment of the reward function with these specific metrics during optimization.
In contrast, MPO relies on preference-based supervision without direct metric-driven rewards.
% \noindent Below are our benchmark results for multiple-choice questions. Looking at the overall data (Tab.~\ref{tab:res_all_choices}), we can see that the Internvl series models perform worse. In the Qwen3-VL series, the model trained using the MPO method performs better, which is attributed to the use of negative samples during training. It also performs well on orthogonal multiple-choice questions, indicating that the model understands changes in viewpoint direction within the video and avoids text hacking. The CRL method model performs well on both multiple-choice questions, further demonstrating the effectiveness of reinforcement learning methods. The VideoNIG-based work proposed in this paper can also explore model performance using various methods.

\noindent In the Choice Evaluation, InternVL series models perform relatively poorly compared to Qwen3-VL variants.
Within the Qwen3-VL family, the MPO-trained model achieves the best overall performance.
This improvement likely stems from preference optimization using diverse negative samples, which sharpens the model's discriminative power under distractor settings.
The MPO-trained model also performs well on Orthogonal-choice, indicating improved robustness to viewpoint shifts and resistance to textual perturbations.
Similarly, the GRPO-trained model demonstrates strong performance across Choice Evaluation, further validating the effectiveness of reinforcement-based optimization for structured navigation reasoning.
Additionally, we observe minimal performance discrepancies between image-conditioned and text-conditioned targets, suggesting that current MLLMs exhibit comparable capability in handling multimodal goal specifications.
% analyse
\noindent Furthermore, we analyze the correlation between four evaluation metrics and Success Rate (SR), a primary performance indicator in VLN.
As shown in \cref{fig:SR_correlation}, although all four metrics exhibit positive correlations with SR, Multiple-choice and Orthogonal-choice show substantially stronger associations compared to Rouge-L and SPICE.
\begin{tcolorbox}[colback=gray!10,colframe=black,arc=3pt,boxrule=0.4pt,left=4pt,right=4pt,top=1pt,bottom=1pt]
\textbf{Takeaway}: Choice Evaluation better reflects practical navigation performance and can provide a more informative signal of instruction quality than traditional text similarity metrics.
\end{tcolorbox}

% \noindent \textbf{Capability of Tour Video Caption.}

% \noindent \textbf{Capability of Localization.}

% \noindent \textbf{Capability of Spatial Reasoning.}
% \vspace{-\baselineskip}
% \vspace{-1em}

% \subsubsection{Generalization of Fundamental VSI Ability}
\begin{table*}[t]
    \centering
    \caption{Navigation execution results for different video types. System2 is InternVLA-N1 + ShortestPathFollower. Dual System is InternVLA-N1 DualVLN.}
    \vspace{-2mm}
    \label{tab:nav_exec}
    \resizebox{\linewidth}{!}{%
    % \frontsize
    \begin{tabular}{l|l|l|cccc|ccccc}
        \toprule[0.8pt]
        \multirow{2}{*}{Video Type} & \multirow{2}{*}{Instruction Type} & \multirow{2}{*}{Model} & \multicolumn{4}{c|}{R2R} & \multicolumn{5}{c}{RxR} \\
        \cmidrule(lr){4-7} \cmidrule(lr){8-12}
        & & & SR $\uparrow$ & SPL $\uparrow$ & OS $\uparrow$ & NE $\downarrow$ & SR $\uparrow$ & SPL $\uparrow$ & OS $\uparrow$ & NE $\downarrow$ & nDTW $\uparrow$ \\
        \midrule[0.5pt]
        \multirow{8}{*}{Gold Route} & \multirow{2}{*}{Ground Truth} & System2 & 58.5 & 53.8 & 65.9 & 4.70 & 55.0 & 47.2 & 64.0 & 5.67 & 66.1 \\
        & & Dual System & \textbf{63.4} & \textbf{57.8} & 69.0 & \textbf{4.29} & \textbf{59.4} & \textbf{50.4} & \textbf{68.2} & \textbf{4.73} & \textbf{69.3} \\
        \cmidrule(lr){2-12}
        & \multirow{2}{*}{Qwen3-VL-8B} & System2 & 40.3 & 34.6 & 50.5 & 6.61 & 29.2 & 23.5 & 40.0 & 8.33 & 48.3 \\
        & & Dual System & 41.6 & 35.4 & 52.4 & 6.21 & 29.2 & 23.5 & 40.0 & 8.33 & 48.3 \\
        \cmidrule(lr){2-12}
        & \multirow{2}{*}{Qwen3-VL-8B(GRPO)} & System2 & 40.1 & 32.2 & 60.4 & 6.57 & 31.4 & 24.6 & 47.8 & 8.43 & 47.3 \\
        & & Dual System & 41.9 & 32.8 & 62.0 & 6.43 & 30.9 & 23.9 & 47.1 & 8.29 & 47.8 \\
        \cmidrule(lr){2-12}
        & \multirow{2}{*}{Qwen3-VL-8B(MPO)} & System2 & 56.1 & 49.8 & 66.7 & 4.84 & 37.5 & 31.6 & 49.7 & 7.23 & 55.3 \\
        & & Dual System & 60.3 & 52.7 & \textbf{69.7} & 4.33 & 40.7 & 33.3 & 51.2 & 6.83 & 56.4 \\
        \midrule[0.5pt]
        \multirow{4}{*}{Tyro Tour} & \multirow{2}{*}{Qwen3-VL-8B} & System2 & 31.0 & 29.2 & 39.0 & 6.42 & 20.4 & 18.7 & 26.6 & 8.87 & 45.6 \\
        & & Dual System & 33.9 & 30.4 & 36.7 & 6.26 & 20.8 & 21.5 & 28.7 & 8.93 & 46.9 \\
        \cmidrule(lr){2-12}
        & \multirow{2}{*}{Qwen3-VL-8B(MPO)} & System2 & 33.2 & 30.5 & \textbf{40.5} & 6.04 & 21.8 & 18.8 & 28.7 & 8.26 & 47.9 \\
        & & Dual System & \textbf{34.3} & \textbf{32.4} & 40.3 & \textbf{6.01} & \textbf{23.4} & \textbf{21.8} & \textbf{29.3} & \textbf{7.95} & \textbf{47.2} \\
        \midrule[0.5pt]
        \multirow{4}{*}{Curiosity Tour} & \multirow{2}{*}{Qwen3-VL-8B} & System2 & 21.0 & 19.2 & 29.0 & 8.22 & 10.4 & 8.77 & 23.9 & 9.87 & 40.6 \\
        & & Dual System & 16.9 & 15.4 & 26.7 & 8.26 & 12.8 & 10.5 & 24.7 & 9.63 & 41.9 \\
        \cmidrule(lr){2-12}
        & \multirow{2}{*}{Qwen3-VL-8B(MPO)} & System2 & \textbf{21.2} & \textbf{19.5} & \textbf{29.9} & \textbf{8.04} & 10.8 & 8.81 & \textbf{24.7} & 9.25 & 40.9 \\
        & & Dual System & 18.2 & 16.4 & 27.3 & 8.21 & \textbf{13.4} & \textbf{11.2} & 21.3 & \textbf{8.55} & \textbf{41.9} \\
        \midrule[0.5pt]
        \multirow{4}{*}{Explorer Tour} & \multirow{2}{*}{Qwen3-VL-8B} & System2 & 18.7 & 17.8 & 25.3 & 8.43 & 8.21 & 8.42 & 23.0 & 9.98 & 38.0 \\
        & & Dual System & 19.3 & 18.2 & 25.9 & \textbf{8.17} & 10.2 & 10.7 & 21.9 & 8.61 & 40.1 \\
        \cmidrule(lr){2-12}
        & \multirow{2}{*}{Qwen3-VL-8B(MPO)} & System2 & \textbf{20.4} & 18.1 & \textbf{25.9} & 8.34 & 10.3 & 8.51 & \textbf{24.2} & 9.75 & 39.2 \\
        & & Dual System & 20.2 & \textbf{18.3} & 25.3 & 8.23 & \textbf{11.4} & \textbf{10.8} & 22.0 & \textbf{8.58} & \textbf{41.2} \\
        \bottomrule[0.8pt]
        % \multicolumn{12}{l}{} \\[0.2em]
        % \multicolumn{12}{l}{Note: System2 is InternVLA-N1 + ShortestPathFollower. Dual System is InternVLA-N1 DualVLN.} \\
    \end{tabular}%
    }
    \vspace{-4mm}
\end{table*}

\noindent \textbf{Generalization of Fundamental VSI Ability. }
% We discuss whether the fundamental ability proved in visual spatial intelligence (VSI) benchmarks is generalizable to our VideoNIG task. The popular VSIBench\cite{yang2025thinking} is selected for comparison, as well as its sub-task, route planning that is related to VideoNIG. As shown in \cref{fig:exp_vsi_nig}, model performance on route planning and VideoNIG generally correlates positively with improvements in fundamental visual spatial understanding.
% As MLLMs continue to grow in capability and increasingly emphasize spatial intelligence, their performance on VideoNIG is expected to improve substantially.
We examine whether visual-spatial abilities measured by VSI-Bench~\cite{yang2025thinking} generalize to VideoNIG.
\cref{fig:exp_vsi_nig} shows positive correlations across metrics.
Pearson $r$ ranges from 0.481 to 0.902, with SPICE and our diagnostic Orthogonal-choice most aligned with general spatial reasoning.
\begin{tcolorbox}[colback=gray!10,colframe=black,arc=3pt,boxrule=0.4pt,left=4pt,right=4pt,top=1pt,bottom=1pt]
\noindent \textbf{Takeaway}: VideoNIG performance strongly tracks visual-spatial reasoning, highlighting it as a key ability for navigation instruction generation.
\end{tcolorbox}

% \begin{wrapfigure}[15]{r}{0.45\linewidth}
%     \centering
%     \vspace{-\baselineskip}
%     \includegraphics[width=\linewidth]{figs/vsi_nig.pdf}
%     \caption{The performance relation between VSIBench, route planning and VideoNIG.}
%     \label{fig:exp_vsi_nig}
% \end{wrapfigure}

% 这里加基于DualVLN的测评结果
% 

% However, the 4B variants of both InternVL3.5 and Qwen3-VL exhibit exceptions.
% Although their overall performance gains, route planning 

% \subsubsection{Factors Influencing Task Difficulty}
% \noindent \textbf{Goal Description Modality.}

% \noindent \textbf{Gold Route Length.}

% \noindent \textbf{Overlap between Gold Route and Tour Videos.}

% \smalltopic{Look-around Actions.}

% \subsubsection{Navigation execution}
\noindent \textbf{Navigation execution.}
% 在基于不同的method上，我们对新的navigation instruction进行了导航执行的测评，我们测试了双系统的性能以及单纯系统2的性能，其中主要的衡量指标是SR/SPL等导航领域的相关指标，从表我们可以看到，相比正确的指令，我们新的生成指令的导航成功率会相对更低，当然这是因为目前大部分的VLN模型都只是在原有的instruction上做了训练，当然这也是目前NIG任务的一个瓶颈，我们无法完全通过VLN的效果来评判Instruction的质量。虽然目前VLN模型在不同指令下的效果还有待提升，但是可以证明从VIdeoNIG-Agent到VLN-Agent的框架是可行的。
% We evaluate navigation execution using both a DualVLN system and a System-2-only setup, reporting SR, SPL, and related metrics.
% Navigation performance under generated instructions is consistently lower than with ground-truth instructions, mainly due to distribution shift, as existing VLN agents are trained on original instruction styles.
% This suggests that end-to-end navigation performance is not a fully reliable proxy for instruction quality. Nevertheless, successful execution indicates the feasibility of the VideoNIG-to-VLN pipeline.
We evaluate navigation with InternVLA-N1~\cite{wei2025ground,internvla-n1} under System2 and Dual System settings, reporting SR, SPL, and related metrics.
Ground-truth instructions perform best because the VLN agent is trained on this distribution.
On Gold Route, generated instructions remain competitive, with MPO consistently outperforming GRPO, likely because the additional supervised fine-tuning stage stabilizes generation.
From Tyro Tour to Curiosity and Explorer Tours, SR and SPL drop under both settings, showing that exploratory trajectories make fine-grained spatial alignment harder.
Although generated instructions still lag behind ground truth under distribution shift, their execution confirms a viable VideoNIG-to-VLN pipeline.

\begin{tcolorbox}[colback=gray!10,colframe=black,arc=3pt,boxrule=0.4pt,left=4pt,right=4pt,top=1pt,bottom=1pt]
    \noindent \textbf{Takeaway}: VideoNIG-generated instructions possess high functional utility for autonomous navigation. 
    However, trajectory complexity remains a key scaling bottleneck, highlighting the challenge of maintaining fine-grained spatial alignment in complex scenarios.
\end{tcolorbox}

\noindent \textbf{Case study. }
\begin{figure}[t]
    \centering
    \includegraphics[width=0.95\linewidth]{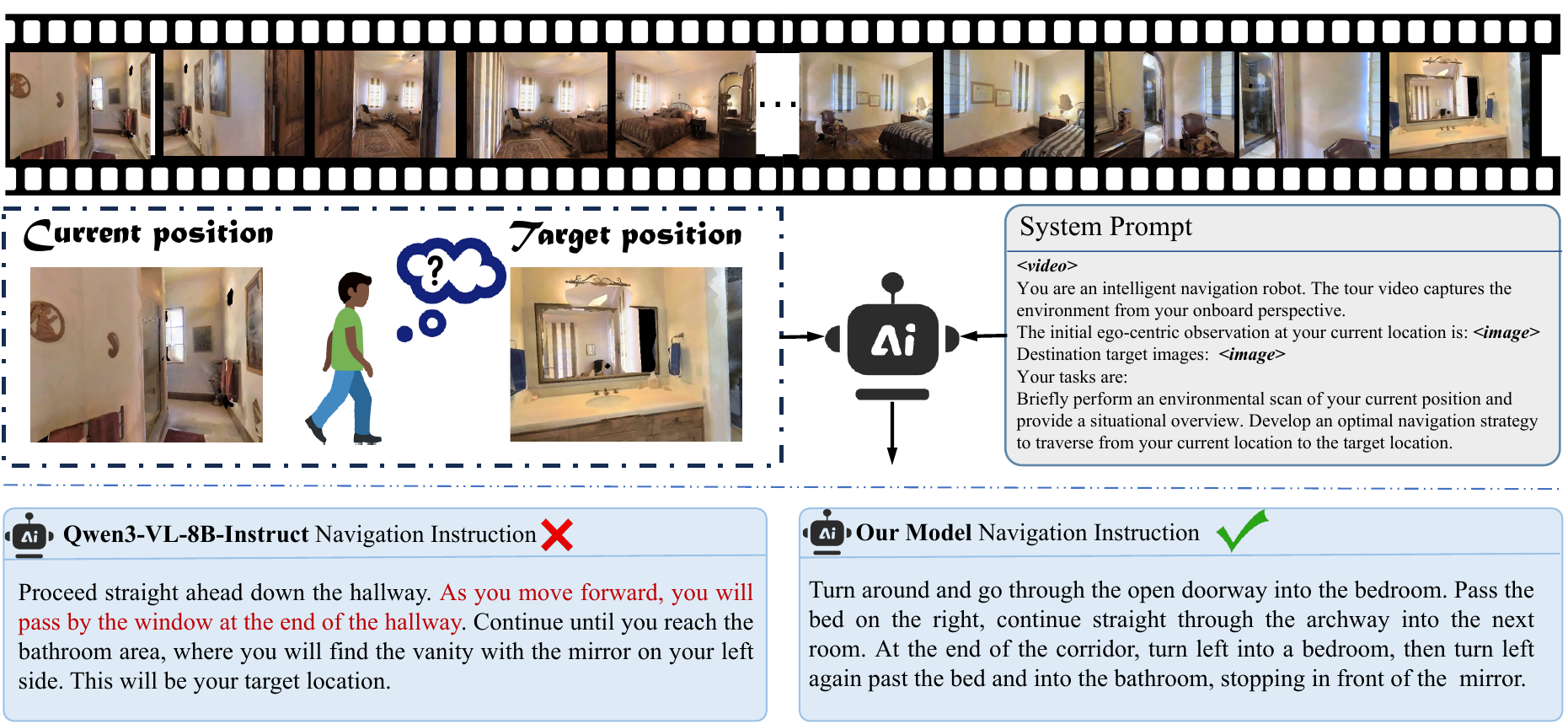}
    \vspace{-2mm}
    \caption{
        Qualitative analysis of a VideoNIG image-goal case.
    }
    \label{fig:case}
    \vspace{-3mm}
\end{figure}
% \noindent \textbf{Case study. }
\cref{fig:case} shows a tour requiring successive turns and consistent orientation tracking.
The Qwen3-VL-8B baseline fails to ground the initial viewpoint and shortcuts down the hallway to the bathroom, while our model follows the multi-turn trajectory, demonstrating improved initial-goal alignment and long-horizon video-grounded reasoning.

\begin{table}[t]
    \centering
    \begin{minipage}[t]{0.48\linewidth}
      \centering
    %   \fontsize{4.5pt}{5.5pt}\selectfont
    %   \renewcommand{\arraystretch}{0.88}
      \setlength{\tabcolsep}{2.5pt}
      \caption{Effects of Chain-of-thought on model performance.  R/S is Rouge-L (top) and SPICE; M/O is Multiple-choice (top) and Orthogonal-choice.}
      \vspace{-2mm}
      \label{tab:ablation_cot}
      \resizebox{\linewidth}{!}{%
      \begin{tabular}{l|cc|cc|cc}
          \toprule[0.6pt]
          Data  & \multicolumn{2}{c}{R2R} & \multicolumn{2}{c}{RxR} & \multicolumn{2}{c}{Average} \\
          \cmidrule(lr){2-3} \cmidrule(lr){4-5} \cmidrule(lr){6-7}
          & R/S & M/O & R/S & M/O & R/S & M/O \\
          \midrule[0.4pt]
          w/ CoT & \makecell{0.301\\0.193} & \makecell{55.28\\56.57} & \makecell{0.256\\0.194} & \makecell{45.94\\52.07} & \makecell{\textbf{0.279}\\\textbf{0.193}} & \makecell{\textbf{50.61}\\\textbf{54.32}}\\
          \midrule[0.4pt]
          w/o CoT & \makecell{0.282\\0.151} & \makecell{55.74\\53.90} & \makecell{0.200\\0.166} & \makecell{44.80\\51.28} & \makecell{0.241\\0.159} & \makecell{50.27\\52.59}\\
          \bottomrule[0.6pt]
      \end{tabular}%
      }
    \end{minipage}\hfill
    \begin{minipage}[t]{0.48\linewidth}
      \centering
    %   \fontsize{4.5pt}{5.5pt}\selectfont
    %   \renewcommand{\arraystretch}{0.88}
    %   \setlength{\tabcolsep}{1.5pt}
    %   \refstepcounter{table}\textbf{Table \thetable:} Curriculum Learning Effects on Model Performance.
    \caption{Curriculum Learning Effects on Model Performance.}
    \vspace{-2mm}
    \label{tab:ablation_cl}
    \resizebox{\linewidth}{!}{%
    \begin{tabular}{c|c|cc|cc|cc}
        \toprule[0.6pt]
        Action & Complexity & \multicolumn{2}{c}{R2R} & \multicolumn{2}{c}{RxR} & \multicolumn{2}{c}{Average} \\
        \cmidrule(lr){3-4} \cmidrule(lr){5-6} \cmidrule(lr){7-8}
        Warmup&Progression & R/S & M/O & R/S & M/O & R/S & M/O \\
        \midrule[0.2pt]
        $\times$ & $\times$ & \makecell{0.283\\0.142} & \makecell{52.65\\58.34} & \makecell{0.148\\0.109} & \makecell{44.01\\46.23} & \makecell{0.216\\0.126} & \makecell{48.33\\52.29}\\
        \midrule[0.2pt]
        \checkmark & $\times$ & \makecell{0.297\\0.143} & \makecell{52.71\\58.44} & \makecell{0.151\\0.113} & \makecell{45.22\\51.01} & \makecell{0.224\\0.128} & \makecell{48.97\\54.73}\\
        \midrule[0.2pt]
        $\times$ & \checkmark & \makecell{0.305\\0.150} & \makecell{60.28\\61.17} & \makecell{0.151\\0.108} & \makecell{45.88\\58.06} & \makecell{0.228\\0.129} & \makecell{53.08\\59.57}\\
        \midrule[0.2pt]
        \checkmark & \checkmark & \makecell{0.334\\0.186} & \makecell{61.27\\66.59} & \makecell{0.191\\0.131} & \makecell{52.39\\59.25} & \makecell{\textbf{0.263}\\\textbf{0.159}} & \makecell{\textbf{56.73}\\\textbf{62.92}}\\
        \bottomrule[0.6pt]
      \end{tabular}%
      }
    \end{minipage}
    \vspace{-4mm}
  \end{table}

\subsection{Ablation Studies}
We ablate Chain-of-thought and Curriculum Learning on VideoNIG.

\noindent \textbf{Effects of Chain-of-thought. }
\cref{tab:ablation_cot} compares the original chain-of-thought prompt, which first analyzes the current state and surroundings before instruction generation, with a direct generation prompt.
Results are reported for the GRPO-based method on unseen Tyro Tour videos with image goals.
% Results include text metrics and multiple-choice test results: R/S: Rouge-L (top)/SPICE (bottom); M/O: Multiple-choice (top) / Orthogonal-choice (bottom). 
The chain-of-thought step lets the model focus on the current environment and state before generation, improving output quality.

\noindent \textbf{Effects of Curriculum Learning.}
% We conduct direction warmup training on Qwen3-VL-8B and evaluate the model on a dedicated directional reasoning benchmark, including action-from-view  and view-from-action.
% After direction warmup training, the model shows substantial improvements in directional reasoning ability as shown in Tab \cref{tab:action_from_view_view_from_action}.
% Furthermore, after full training with the MPO, the performance on the VideoNIG task is further improved, as shown in \cref{tab:directional_pre_training_ablation}.
\cref{tab:ablation_cl} evaluates the MPO-trained model on unseen Tyro Tour videos with image goals.
Without Curriculum Learning, performance is consistently lower, indicating unstable optimization for long-horizon objectives.
Action Warmup alleviates this by restricting early-stage optimization to relative spatial reasoning, providing a more consistent training signal and improving directional discrimination.
Building on this foundation, Complexity Progression gradually increases route length and difficulty, allowing the model to adapt to harder scenarios while preserving spatial reasoning.
Overall, the two-stage curriculum stabilizes optimization and yields consistent gains across metrics.

\section{Conclusion}
% We introduce \textbf{VideoNIG}, a novel goal-oriented navigation instruction generation task that leverages ego-centric tour videos as compact environmental priors.
% Building on R2R-CE and RxR-CE in continuous environments, we instantiate a controlled benchmark comprising tours generated by different methods and multimodal goal descriptions. Our experiments demonstrate that state-of-the-art MLLMs still find navigation instruction generation from tour videos challenging.
% To better activate visual-spatial capabilities, we propose a two-stage Curriculum Learning method that effectively improves both instruction quality and downstream VLN performance.

% \noindent Our work has limitations: the dataset is fully simulator-based and restricted to indoor environments, and evaluation relies mainly on automatic NLP metrics. 
% In future work, we plan to extend VideoNIG to physical and outdoor scenarios with robot-collected tours, explore broader backbones and richer multimodal priors (e.g., combining videos with maps or 3D reconstructions), and design human-aligned, interaction-aware settings that allow agents to iteratively refine instructions and update environmental priors.
We introduce \textbf{VideoNIG}, a goal-oriented navigation instruction generation task that uses ego-centric tour videos as compact environmental priors.
Built on R2R-CE and RxR-CE, our controlled benchmark instantiation provides diverse tour strategies and multimodal goals for studying long-horizon video-grounded instruction generation.
Results show that strong MLLMs still struggle with spatial grounding and trajectory composition, while our two-stage Curriculum Learning framework improves both instruction quality and downstream VLN performance.
Although limited to simulator-based indoor settings, VideoNIG offers a compact evaluation setting for embodied multimodal reasoning and motivates richer multimodal priors and interaction-aware scenarios.

% \clearpage  % TODO FINAL: Remove for camera-ready if references fit on same page

% \section*{Acknowledgements}
% Please insert your acknowledgments here.

\bibliographystyle{splncs04}
\bibliography{main}

@article{wei2025streamvln,
  author = {Wei, Meng and Wan, Chenyang and Yu, Xiqian and Wang, Tai and Yang, Yuqiang and Mao, Xiaohan and Zhu, Chenming and Cai, Wenzhe and Wang, Hanqing and Chen, Yilun and others},
  journal = {arXiv preprint arXiv:2507.05240},
  year = {2025},
  title = {Streamvln: Streaming vision-and-language navigation via slowfast context modeling},
}

@article{zhang2024navid,
  author = {Zhang, Jiazhao and Wang, Kunyu and Xu, Rongtao and Zhou, Gengze and Hong, Yicong and Fang, Xiaomeng and Wu, Qi and Zhang, Zhizheng and Wang, He},
  journal = {arXiv preprint arXiv:2402.15852},
  year = {2024},
  title = {Navid: Video-based vlm plans the next step for vision-and-language navigation},
}

@inproceedings{anderson2018r2r,
  author = {Anderson, Peter and Wu, Qi and Teney, Damien and Bruce, Jake and Johnson, Mark and S{\"u}nderhauf, Niko and Reid, Ian and Gould, Stephen and Van Den Hengel, Anton},
  booktitle = {Proceedings of the IEEE conference on computer vision and pattern recognition},
  pages = {3674--3683},
  year = {2018},
  title = {Vision-and-language navigation: Interpreting visually-grounded navigation instructions in real environments},
}

@article{ku2020rxr,
  author = {Ku, Alexander and Anderson, Peter and Patel, Roma and Ie, Eugene and Baldridge, Jason},
  journal = {arXiv preprint arXiv:2010.07954},
  year = {2020},
  title = {Room-across-room: Multilingual vision-and-language navigation with dense spatiotemporal grounding},
}

@inproceedings{krantz2020vlnce,
  author = {Krantz, Jacob and Wijmans, Erik and Majumdar, Arjun and Batra, Dhruv and Lee, Stefan},
  booktitle = {European Conference on Computer Vision},
  organization = {Springer},
  pages = {104--120},
  year = {2020},
  title = {Beyond the nav-graph: Vision-and-language navigation in continuous environments},
}

@inproceedings{zhou2024navgpt,
  author = {Zhou, Gengze and Hong, Yicong and Wang, Zun and Wang, Xin Eric and Wu, Qi},
  booktitle = {European Conference on Computer Vision},
  organization = {Springer},
  pages = {260--278},
  year = {2024},
  title = {Navgpt-2: Unleashing navigational reasoning capability for large vision-language models},
}

@article{an2024etpnav,
  author = {An, Dong and Wang, Hanqing and Wang, Wenguan and Wang, Zun and Huang, Yan and He, Keji and Wang, Liang},
  journal = {IEEE Transactions on Pattern Analysis and Machine Intelligence},
  year = {2024},
  publisher = {IEEE},
  title = {Etpnav: Evolving topological planning for vision-language navigation in continuous environments},
}

@article{hong2025general,
  author = {Hong, Haodong and Qiao, Yanyuan and Wang, Sen and Liu, Jiajun and Wu, Qi},
  journal = {arXiv preprint arXiv:2501.17403},
  year = {2025},
  title = {General Scene Adaptation for Vision-and-Language Navigation},
}

@article{li2025ground,
  author = {Li, Zerui and Zhou, Gengze and Hong, Haodong and Shao, Yanyan and Lyu, Wenqi and Qiao, Yanyuan and Wu, Qi},
  journal = {arXiv preprint arXiv:2502.19024},
  year = {2025},
  title = {Ground-level viewpoint vision-and-language navigation in continuous environments},
}

@article{hirose2024lelan,
  author = {Hirose, Noriaki and Glossop, Catherine and Sridhar, Ajay and Shah, Dhruv and Mees, Oier and Levine, Sergey},
  journal = {arXiv preprint arXiv:2410.03603},
  year = {2024},
  title = {Lelan: Learning a language-conditioned navigation policy from in-the-wild videos},
}

@inproceedings{shah2023lm,
  author = {Shah, Dhruv and Osi{\'n}ski, B{\l}a{\.z}ej and Levine, Sergey and others},
  booktitle = {Conference on robot learning},
  organization = {PMLR},
  pages = {492--504},
  year = {2023},
  title = {Lm-nav: Robotic navigation with large pre-trained models of language, vision, and action},
}

@inproceedings{zheng2024navillm,
  author = {Zheng, Duo and Huang, Shijia and Zhao, Lin and Zhong, Yiwu and Wang, Liwei},
  booktitle = {Proceedings of the IEEE/CVF Conference on Computer Vision and Pattern Recognition},
  pages = {13624--13634},
  year = {2024},
  title = {Towards learning a generalist model for embodied navigation},
}

@inproceedings{song2025towards,
  author = {Song, Xinshuai and Chen, Weixing and Liu, Yang and Chen, Weikai and Li, Guanbin and Lin, Liang},
  booktitle = {Proceedings of the Computer Vision and Pattern Recognition Conference},
  pages = {12078--12088},
  year = {2025},
  title = {Towards long-horizon vision-language navigation: Platform, benchmark and method},
}

@article{chiang2024mobility,
  author = {Chiang, Hao-Tien Lewis and Xu, Zhuo and Fu, Zipeng and Jacob, Mithun George and Zhang, Tingnan and Lee, Tsang-Wei Edward and Yu, Wenhao and Schenck, Connor and Rendleman, David and Shah, Dhruv and others},
  journal = {arXiv preprint arXiv:2407.07775},
  year = {2024},
  title = {Mobility vla: Multimodal instruction navigation with long-context vlms and topological graphs},
}

@article{cheng2024navila,
  author = {Cheng, An-Chieh and Ji, Yandong and Yang, Zhaojing and Gongye, Zaitian and Zou, Xueyan and Kautz, Jan and B{\i}y{\i}k, Erdem and Yin, Hongxu and Liu, Sifei and Wang, Xiaolong},
  journal = {arXiv preprint arXiv:2412.04453},
  year = {2024},
  title = {Navila: Legged robot vision-language-action model for navigation},
}

@article{zhang2024uninavid,
  author = {Zhang, Jiazhao and Wang, Kunyu and Wang, Shaoan and Li, Minghan and Liu, Haoran and Wei, Songlin and Wang, Zhongyuan and Zhang, Zhizheng and Wang, He},
  journal = {arXiv preprint arXiv:2412.06224},
  year = {2024},
  title = {Uni-navid: A video-based vision-language-action model for unifying embodied navigation tasks},
}

@article{long2024instructnav,
  author = {Long, Yuxing and Cai, Wenzhe and Wang, Hongcheng and Zhan, Guanqi and Dong, Hao},
  journal = {arXiv preprint arXiv:2406.04882},
  year = {2024},
  title = {Instructnav: Zero-shot system for generic instruction navigation in unexplored environment},
}

@article{an2022bevbert,
  author = {An, Dong and Qi, Yuankai and Li, Yangguang and Huang, Yan and Wang, Liang and Tan, Tieniu and Shao, Jing},
  journal = {arXiv preprint arXiv:2212.04385},
  year = {2022},
  title = {Bevbert: Multimodal map pre-training for language-guided navigation},
}

@inproceedings{wijmans2019embodied,
  author = {Wijmans, Erik and Datta, Samyak and Maksymets, Oleksandr and Das, Abhishek and Gkioxari, Georgia and Lee, Stefan and Essa, Irfan and Parikh, Devi and Batra, Dhruv},
  booktitle = {Proceedings of the IEEE/CVF Conference on Computer Vision and Pattern Recognition},
  pages = {6659--6668},
  year = {2019},
  title = {Embodied question answering in photorealistic environments with point cloud perception},
}

@inproceedings{yokoyama2024hm3d,
  author = {Yokoyama, Naoki and Ramrakhya, Ram and Das, Abhishek and Batra, Dhruv and Ha, Sehoon},
  booktitle = {2024 IEEE/RSJ International Conference on Intelligent Robots and Systems (IROS)},
  organization = {IEEE},
  pages = {5543--5550},
  year = {2024},
  title = {Hm3d-ovon: A dataset and benchmark for open-vocabulary object goal navigation},
}

@inproceedings{zhu2021soon,
  author = {Zhu, Fengda and Liang, Xiwen and Zhu, Yi and Yu, Qizhi and Chang, Xiaojun and Liang, Xiaodan},
  booktitle = {Proceedings of the IEEE/CVF Conference on Computer Vision and Pattern Recognition},
  pages = {12689--12699},
  year = {2021},
  title = {Soon: Scenario oriented object navigation with graph-based exploration},
}

@article{zhang2025mapnav,
  author = {Zhang, L and Hao, X and Xu, Q and Zhang, Q and Zhang, X and Wang, P and Zhang, J and Wang, Z and Zhang, S and Xu, R MapNav},
  journal = {arXiv preprint arXiv:2502.13451},
  year = {2025},
  title = {A novel memory representation via annotated semantic maps for vlm-based vision-and-language navigation},
}

@inproceedings{han2025dialnav,
  author = {Han, Leekyeung and Min, Hyunji and Hwangbo, Gyeom and Choi, Jonghyun and Seo, Paul Hongsuck},
  booktitle = {Proceedings of the IEEE/CVF International Conference on Computer Vision},
  pages = {8514--8523},
  year = {2025},
  title = {DialNav: Multi-turn Dialog Navigation with a Remote Guide},
}

@article{zhang2025mem2ego,
  author = {Zhang, Lingfeng and Liu, Yuecheng and Zhang, Zhanguang and Aghaei, Matin and Hu, Yaochen and Gu, Hongjian and Alomrani, Mohammad Ali and Bravo, David Gamaliel Arcos and Karimi, Raika and Hamidizadeh, Atia and others},
  journal = {arXiv preprint arXiv:2502.14254},
  year = {2025},
  title = {Mem2ego: Empowering vision-language models with global-to-ego memory for long-horizon embodied navigation},
}

@article{fried2018speaker,
  author = {Fried, Daniel and Hu, Ronghang and Cirik, Volkan and Rohrbach, Anna and Andreas, Jacob and Morency, Louis-Philippe and Berg-Kirkpatrick, Taylor and Saenko, Kate and Klein, Dan and Darrell, Trevor},
  journal = {Advances in neural information processing systems},
  year = {2018},
  title = {Speaker-follower models for vision-and-language navigation},
  volume = {31}
}

@inproceedings{qi2020reverie,
  author = {Qi, Yuankai and Wu, Qi and Anderson, Peter and Wang, Xin and Wang, William Yang and Shen, Chunhua and Hengel, Anton van den},
  booktitle = {Proceedings of the IEEE/CVF Conference on Computer Vision and Pattern Recognition},
  pages = {9982--9991},
  year = {2020},
  title = {Reverie: Remote embodied visual referring expression in real indoor environments},
}

@article{ramrakhya2025grounding,
  author = {Ramrakhya, Ram and Chang, Matthew and Puig, Xavier and Desai, Ruta and Kira, Zsolt and Mottaghi, Roozbeh},
  journal = {arXiv preprint arXiv:2504.00907},
  year = {2025},
  title = {Grounding Multimodal LLMs to Embodied Agents that Ask for Help with Reinforcement Learning},
}

@inproceedings{anwar2025remembr,
  author = {Anwar, Abrar and Welsh, John and Biswas, Joydeep and Pouya, Soha and Chang, Yan},
  booktitle = {2025 IEEE International Conference on Robotics and Automation (ICRA)},
  organization = {IEEE},
  pages = {2838--2845},
  year = {2025},
  title = {Remembr: Building and reasoning over long-horizon spatio-temporal memory for robot navigation},
}

@article{tan2019learning,
  author = {Tan, Hao and Yu, Licheng and Bansal, Mohit},
  journal = {arXiv preprint arXiv:1904.04195},
  year = {2019},
  title = {Learning to navigate unseen environments: Back translation with environmental dropout},
}

@inproceedings{han2025roomtour3d,
  author = {Han, Mingfei and Ma, Liang and Zhumakhanova, Kamila and Radionova, Ekaterina and Zhang, Jingyi and Chang, Xiaojun and Liang, Xiaodan and Laptev, Ivan},
  booktitle = {Proceedings of the Computer Vision and Pattern Recognition Conference},
  pages = {27586--27596},
  year = {2025},
  title = {Roomtour3d: Geometry-aware video-instruction tuning for embodied navigation},
}

@inproceedings{chen2022think,
  author = {Chen, Shizhe and Guhur, Pierre-Louis and Tapaswi, Makarand and Schmid, Cordelia and Laptev, Ivan},
  booktitle = {Proceedings of the IEEE/CVF Conference on Computer Vision and Pattern Recognition},
  pages = {16537--16547},
  year = {2022},
  title = {Think global, act local: Dual-scale graph transformer for vision-and-language navigation},
}

@article{wu2025govig,
  author = {Wu, Fengyi and Dong, Yifei and Cheng, Zhi-Qi and Dai, Yilong and Chen, Guangyu and Wang, Hang and Dai, Qi and Hauptmann, Alexander G},
  journal = {arXiv preprint arXiv:2508.09547},
  year = {2025},
  title = {Govig: Goal-conditioned visual navigation instruction generation},
}

@inproceedings{wang2023lana,
  author = {Wang, Xiaohan and Wang, Wenguan and Shao, Jiayi and Yang, Yi},
  booktitle = {Proceedings of the IEEE/CVF conference on computer vision and pattern recognition},
  pages = {19048--19058},
  year = {2023},
  title = {Lana: A language-capable navigator for instruction following and generation},
}

@inproceedings{lin2023learning,
  author = {Lin, Kunyang and Chen, Peihao and Huang, Diwei and Li, Thomas H and Tan, Mingkui and Gan, Chuang},
  booktitle = {Proceedings of the IEEE/CVF International Conference on Computer Vision},
  pages = {8317--8326},
  year = {2023},
  title = {Learning vision-and-language navigation from youtube videos},
}

@inproceedings{wang2022less,
  author = {Wang, Su and Montgomery, Ceslee and Orbay, Jordi and Birodkar, Vighnesh and Faust, Aleksandra and Gur, Izzeddin and Jaques, Natasha and Waters, Austin and Baldridge, Jason and Anderson, Peter},
  booktitle = {Proceedings of the IEEE/CVF Conference on Computer Vision and Pattern Recognition},
  pages = {15428--15438},
  year = {2022},
  title = {Less is more: Generating grounded navigation instructions from landmarks},
}

@inproceedings{wang2022counterfactual,
  author = {Wang, Hanqing and Liang, Wei and Shen, Jianbing and Van Gool, Luc and Wang, Wenguan},
  booktitle = {Proceedings of the IEEE/CVF conference on computer vision and pattern recognition},
  pages = {15471--15481},
  year = {2022},
  title = {Counterfactual cycle-consistent learning for instruction following and generation in vision-language navigation},
}

@article{wang2025navrag,
  author = {Wang, Zihan and Zhu, Yaohui and Lee, Gim Hee and Fan, Yachun},
  journal = {arXiv preprint arXiv:2502.11142},
  year = {2025},
  title = {Navrag: Generating user demand instructions for embodied navigation through retrieval-augmented llm},
}

@article{wang2024bootstrapping,
  author = {Wang, Zun and Li, Jialu and Hong, Yicong and Li, Songze and Li, Kunchang and Yu, Shoubin and Wang, Yi and Qiao, Yu and Wang, Yali and Bansal, Mohit and others},
  journal = {arXiv preprint arXiv:2412.08467},
  year = {2024},
  title = {Bootstrapping language-guided navigation learning with self-refining data flywheel},
}

@inproceedings{wang2023scaling,
  author = {Wang, Zun and Li, Jialu and Hong, Yicong and Wang, Yi and Wu, Qi and Bansal, Mohit and Gould, Stephen and Tan, Hao and Qiao, Yu},
  booktitle = {Proceedings of the IEEE/CVF international conference on computer vision},
  pages = {12009--12020},
  year = {2023},
  title = {Scaling data generation in vision-and-language navigation},
}

@inproceedings{fan2024navigation,
  author = {Fan, Sheng and Liu, Rui and Wang, Wenguan and Yang, Yi},
  booktitle = {European Conference on Computer Vision},
  organization = {Springer},
  pages = {368--387},
  year = {2024},
  title = {Navigation instruction generation with bev perception and large language models},
}

@inproceedings{fan2025scene,
  author = {Fan, Sheng and Liu, Rui and Wang, Wenguan and Yang, Yi},
  booktitle = {Proceedings of the Computer Vision and Pattern Recognition Conference},
  pages = {6898--6908},
  year = {2025},
  title = {Scene map-based prompt tuning for navigation instruction generation},
}

@article{cui2025generating,
  author = {Cui, Yibo and Xie, Liang and Zhao, Yu and Sun, Jiawei and Yin, Erwei},
  journal = {arXiv preprint arXiv:2506.08566},
  year = {2025},
  title = {Generating Vision-Language Navigation Instructions Incorporated Fine-Grained Alignment Annotations},
}

@inproceedings{yang2025thinking,
  author = {Yang, Jihan and Yang, Shusheng and Gupta, Anjali W and Han, Rilyn and Fei-Fei, Li and Xie, Saining},
  booktitle = {Proceedings of the Computer Vision and Pattern Recognition Conference},
  pages = {10632--10643},
  year = {2025},
  title = {Thinking in space: How multimodal large language models see, remember, and recall spaces},
}

@inproceedings{zhou2025vlm4d,
  author = {Zhou, Shijie and Vilesov, Alexander and He, Xuehai and Wan, Ziyu and Zhang, Shuwang and Nagachandra, Aditya and Chang, Di and Chen, Dongdong and Wang, Xin Eric and Kadambi, Achuta},
  booktitle = {Proceedings of the IEEE/CVF international conference on computer vision},
  pages = {8600--8612},
  year = {2025},
  title = {Vlm4d: Towards spatiotemporal awareness in vision language models},
}

@inproceedings{wang2025vggt,
  author = {Wang, Jianyuan and Chen, Minghao and Karaev, Nikita and Vedaldi, Andrea and Rupprecht, Christian and Novotny, David},
  booktitle = {Proceedings of the Computer Vision and Pattern Recognition Conference},
  pages = {5294--5306},
  year = {2025},
  title = {Vggt: Visual geometry grounded transformer},
}

@inproceedings{bigverdi2025perception,
  author = {Bigverdi, Mahtab and Luo, Zelun and Hsieh, Cheng-Yu and Shen, Ethan and Chen, Dongping and Shapiro, Linda G and Krishna, Ranjay},
  booktitle = {Proceedings of the Computer Vision and Pattern Recognition Conference},
  pages = {3836--3845},
  year = {2025},
  title = {Perception tokens enhance visual reasoning in multimodal language models},
}

@article{huang2025mllms,
  author = {Huang, Xiaohu and Wu, Jingjing and Xie, Qunyi and Han, Kai},
  journal = {arXiv preprint arXiv:2506.01946},
  year = {2025},
  title = {MLLMs Need 3D-Aware Representation Supervision for Scene Understanding},
}

@article{wang2025internvl35,
  author = {Wang, Weiyun and Gao, Zhangwei and Gu, Lixin and Pu, Hengjun and Cui, Long and Wei, Xingguang and Liu, Zhaoyang and Jing, Linglin and Ye, Shenglong and Shao, Jie and others},
  journal = {arXiv preprint arXiv:2508.18265},
  year = {2025},
  title = {Internvl3. 5: Advancing open-source multimodal models in versatility, reasoning, and efficiency},
}

@misc{bai2025qwen3vl,
    title        = {Qwen3-VL Github Repo.},
    author       = {Bai, Shuai and others},
    year         = {2025}
}

@inproceedings{wang2025himtok,
  author = {Wang, Tao and Cheng, Changxu and Wang, Lingfeng and Chen, Senda and Zhao, Wuyue},
  booktitle = {Proceedings of the IEEE/CVF International Conference on Computer Vision},
  pages = {23267--23278},
  year = {2025},
  title = {Himtok: Learning hierarchical mask tokens for image segmentation with large multimodal model},
}

@article{zhang2024vision,
  author = {Zhang, Yue and Ma, Ziqiao and Li, Jialu and Qiao, Yanyuan and Wang, Zun and Chai, Joyce and Wu, Qi and Bansal, Mohit and Kordjamshidi, Parisa},
  journal = {arXiv preprint arXiv:2407.07035},
  year = {2024},
  title = {Vision-and-language navigation today and tomorrow: A survey in the era of foundation models},
}

@article{zhao2021evaluation,
  author = {Zhao, Ming and Anderson, Peter and Jain, Vihan and Wang, Su and Ku, Alexander and Baldridge, Jason and Ie, Eugene},
  journal = {arXiv preprint arXiv:2101.10504},
  year = {2021},
  title = {On the evaluation of vision-and-language navigation instructions},
}

@inproceedings{huang2022assister,
  author = {Huang, Zanming and Shangguan, Zhongkai and Zhang, Jimuyang and Bar, Gilad and Boyd, Matthew and Ohn-Bar, Eshed},
  booktitle = {European Conference on Computer Vision},
  organization = {Springer},
  pages = {271--289},
  year = {2022},
  title = {Assister: Assistive navigation via conditional instruction generation},
}

@article{dorbala2024can,
  author = {Dorbala, Vishnu Sashank and Chowdhury, Sanjoy and Manocha, Dinesh},
  journal = {arXiv preprint arXiv:2403.11487},
  year = {2024},
  title = {Can llms generate human-like wayfinding instructions? towards platform-agnostic embodied instruction synthesis},
}

@inproceedings{yokoyama2024vlfm,
  author = {Yokoyama, Naoki and Ha, Sehoon and Batra, Dhruv and Wang, Jiuguang and Bucher, Bernadette},
  booktitle = {2024 IEEE International Conference on Robotics and Automation (ICRA)},
  organization = {IEEE},
  pages = {42--48},
  year = {2024},
  title = {Vlfm: Vision-language frontier maps for zero-shot semantic navigation},
}

@inproceedings{chen2025moto,
  author = {Chen, Yi and Ge, Yuying and Tang, Weiliang and Li, Yizhuo and Ge, Yixiao and Ding, Mingyu and Shan, Ying and Liu, Xihui},
  booktitle = {Proceedings of the IEEE/CVF International Conference on Computer Vision},
  pages = {19752--19763},
  year = {2025},
  title = {Moto: Latent motion token as the bridging language for learning robot manipulation from videos},
}

@inproceedings{savva2019habitat,
  author = {Savva, Manolis and Kadian, Abhishek and Maksymets, Oleksandr and Zhao, Yili and Wijmans, Erik and Jain, Bhavana and Straub, Julian and Liu, Jia and Koltun, Vladlen and Malik, Jitendra and others},
  booktitle = {Proceedings of the IEEE/CVF international conference on computer vision},
  pages = {9339--9347},
  year = {2019},
  title = {Habitat: A platform for embodied ai research},
}

@article{ilharco2019ndtw,
  author = {Ilharco, Gabriel and Jain, Vihan and Ku, Alexander and Ie, Eugene and Baldridge, Jason},
  journal = {arXiv preprint arXiv:1907.05446},
  year = {2019},
  title = {General evaluation for instruction conditioned navigation using dynamic time warping},
}

@article{fischler1981random,
  author = {Fischler, Martin A and Bolles, Robert C},
  journal = {Communications of the ACM},
  number = {6},
  pages = {381--395},
  year = {1981},
  publisher = {ACM New York, NY, USA},
  title = {Random sample consensus: a paradigm for model fitting with applications to image analysis and automated cartography},
  volume = {24}
}

@inproceedings{liu2008isolation,
  author = {Liu, Fei Tony and Ting, Kai Ming and Zhou, Zhi-Hua},
  booktitle = {2008 eighth ieee international conference on data mining},
  organization = {IEEE},
  pages = {413--422},
  year = {2008},
  title = {Isolation forest},
}

@article{zhu2025internvl3,
  author = {Zhu, Jinguo and Wang, Weiyun and Chen, Zhe and Liu, Zhaoyang and Ye, Shenglong and Gu, Lixin and Tian, Hao and Duan, Yuchen and Su, Weijie and Shao, Jie and others},
  journal = {arXiv preprint arXiv:2504.10479},
  year = {2025},
  title = {Internvl3: Exploring advanced training and test-time recipes for open-source multimodal models},
}

@article{bai2025qwen2_5,
  author = {Bai, Shuai and Chen, Keqin and Liu, Xuejing and Wang, Jialin and Ge, Wenbin and Song, Sibo and Dang, Kai and Wang, Peng and Wang, Shijie and Tang, Jun and others},
  journal = {arXiv preprint arXiv:2502.13923},
  year = {2025},
  title = {Qwen2. 5-vl technical report},
}

@inproceedings{anderson2016spice,
  author = {Anderson, Peter and Fernando, Basura and Johnson, Mark and Gould, Stephen},
  booktitle = {European conference on computer vision},
  organization = {Springer},
  pages = {382--398},
  year = {2016},
  title = {Spice: Semantic propositional image caption evaluation},
}

@inproceedings{lin2004rouge,
  author = {Lin, Chin-Yew},
  booktitle = {Text summarization branches out},
  pages = {74--81},
  year = {2004},
  title = {Rouge: A package for automatic evaluation of summaries},
}

@article{chen2024allava,
  author = {Chen, Guiming Hardy and Chen, Shunian and Zhang, Ruifei and Chen, Junying and Wu, Xiangbo and Zhang, Zhiyi and Chen, Zhihong and Li, Jianquan and Wan, Xiang and Wang, Benyou},
  journal = {arXiv preprint arXiv:2402.11684},
  year = {2024},
  title = {Allava: Harnessing gpt4v-synthesized data for lite vision-language models},
}

@article{conover2023free,
  author = {Conover, Mike and Hayes, Matt and Mathur, Ankit and Xie, Jianwei and Wan, Jun and Shah, Sam and Ghodsi, Ali and Wendell, Patrick and Zaharia, Matei and Xin, Reynold},
  year = {2023},
  title = {Free dolly: Introducing the world’s first truly open instructiontuned llm},
}

@article{zheng2024opencodeinterpreter,
  author = {Zheng, Tianyu and Zhang, Ge and Shen, Tianhao and Liu, Xueling and Lin, Bill Yuchen and Fu, Jie and Chen, Wenhu and Yue, Xiang},
  journal = {arXiv preprint arXiv:2402.14658},
  year = {2024},
  title = {Opencodeinterpreter: Integrating code generation with execution and refinement},
}

@article{yue2023mammoth,
  author = {Yue, Xiang and Qu, Xingwei and Zhang, Ge and Fu, Yao and Huang, Wenhao and Sun, Huan and Su, Yu and Chen, Wenhu},
  journal = {arXiv preprint arXiv:2309.05653},
  year = {2023},
  title = {Mammoth: Building math generalist models through hybrid instruction tuning},
}

@article{yu2023metamath,
  author = {Yu, Longhui and Jiang, Weisen and Shi, Han and Yu, Jincheng and Liu, Zhengying and Zhang, Yu and Kwok, James T and Li, Zhenguo and Weller, Adrian and Liu, Weiyang},
  journal = {arXiv preprint arXiv:2309.12284},
  year = {2023},
  title = {Metamath: Bootstrap your own mathematical questions for large language models},
}

@article{mitra2024orca,
  author = {Mitra, Arindam and Khanpour, Hamed and Rosset, Corby and Awadallah, Ahmed},
  journal = {arXiv preprint arXiv:2402.14830},
  year = {2024},
  title = {Orca-math: Unlocking the potential of slms in grade school math},
}

@article{ouyang2025spacer,
  author = {Ouyang, Kun and Liu, Yuanxin and Wu, Haoning and Liu, Yi and Zhou, Hao and Zhou, Jie and Meng, Fandong and Sun, Xu},
  journal = {arXiv preprint arXiv:2504.01805},
  year = {2025},
  title = {SpaceR: Reinforcing MLLMs in Video Spatial Reasoning},
}

@article{wang2024cogvlm,
  author = {Wang, Weihan and Lv, Qingsong and Yu, Wenmeng and Hong, Wenyi and Qi, Ji and Wang, Yan and Ji, Junhui and Yang, Zhuoyi and Zhao, Lei and XiXuan, Song and others},
  journal = {Advances in Neural Information Processing Systems},
  pages = {121475--121499},
  year = {2024},
  title = {Cogvlm: Visual expert for pretrained language models},
  volume = {37}
}

@inproceedings{bengio2009curriculum,
  title={Curriculum learning},
  author={Bengio, Yoshua and Louradour, J{\'e}r{\^o}me and Collobert, Ronan and Weston, Jason},
  booktitle={Proceedings of the 26th annual international conference on machine learning},
  pages={41--48},
  year={2009}
}

@article{shao2024deepseekmath,
  title={Deepseekmath: Pushing the limits of mathematical reasoning in open language models},
  author={Shao, Zhihong and Wang, Peiyi and Zhu, Qihao and Xu, Runxin and Song, Junxiao and Bi, Xiao and Zhang, Haowei and Zhang, Mingchuan and Li, YK and Wu, Yang and others},
  journal={arXiv preprint arXiv:2402.03300},
  year={2024}
}

@article{qiao2025navbench,
  title={NavBench: Probing Multimodal Large Language Models for Embodied Navigation},
  author={Qiao, Yanyuan and Hong, Haodong and Lyu, Wenqi and An, Dong and Zhang, Siqi and Xie, Yutong and Wang, Xinyu and Wu, Qi},
  journal={arXiv preprint arXiv:2506.01031},
  year={2025}
}

@article{wei2025ground,
  title={Ground slow, move fast: A dual-system foundation model for generalizable vision-and-language navigation},
  author={Wei, Meng and Wan, Chenyang and Peng, Jiaqi and Yu, Xiqian and Yang, Yuqiang and Feng, Delin and Cai, Wenzhe and Zhu, Chenming and Wang, Tai and Pang, Jiangmiao and others},
  journal={arXiv preprint arXiv:2512.08186},
  year={2025}
}

@article{wang2024enhancing,
  title={Enhancing the reasoning ability of multimodal large language models via mixed preference optimization},
  author={Wang, Weiyun and Chen, Zhe and Wang, Wenhai and Cao, Yue and Liu, Yangzhou and Gao, Zhangwei and Zhu, Jinguo and Zhu, Xizhou and Lu, Lewei and Qiao, Yu and others},
  journal={arXiv preprint arXiv:2411.10442},
  year={2024}
}

@inproceedings{kwon2023efficient,
  title={Efficient Memory Management for Large Language Model Serving with PagedAttention},
  author={Woosuk Kwon and Zhuohan Li and Siyuan Zhuang and Ying Sheng and Lianmin Zheng and Cody Hao Yu and Joseph E. Gonzalez and Hao Zhang and Ion Stoica},
  booktitle={Proceedings of the ACM SIGOPS 29th Symposium on Operating Systems Principles},
  year={2023}
}

@inproceedings{liu2025lamra,
  title={Lamra: Large multimodal model as your advanced retrieval assistant},
  author={Liu, Yikun and Zhang, Yajie and Cai, Jiayin and Jiang, Xiaolong and Hu, Yao and Yao, Jiangchao and Wang, Yanfeng and Xie, Weidi},
  booktitle={Proceedings of the Computer Vision and Pattern Recognition Conference},
  pages={4015--4025},
  year={2025}
}

@misc{internvla-n1,
    title = {{InternVLA-N1: An} Open Dual-System Navigation Foundation Model with Learned Latent Plans},
    author = {InternNav Team},
    year = {2025},
    booktitle={arXiv},
}
% Supplementary materials
\clearpage
\setcounter{page}{1}
% \maketitlesupplementary
\appendix

% \section{Limitations}
% \label{limit}
% static, lack of exploration and environment updation, rely on static priors
% not combined with VLN agent

% \section{Datails on VideoNIG Dataset}

% \label{data_details}

% \title{Appendix}
\section{Details on VideoNIG Dataset}
\label{sec: tour_details}

In this section, we provide the detailed pipeline for constructing the three types of tour videos used in VideoNIG: \textit{Tyro Tour}, \textit{Curiosity Tour}, and \textit{Explorer Tour}. 
All tours are generated in the Habitat simulator, following the discretized action space \{turn-left, turn-right, move-forward\}. 
For each action taken, one frame is captured at 6 FPS.

\subsection{Overview}
To ensure that all tours retain the essential visual coverage of the underlying optimal path, we use the Gold Route---the shortest path between the annotated start and goal---as the structural reference. Diversity is introduced by (1) sampling randomized start/end points, (2) inserting look-around behaviors along the route, and (3) performing neighborhood exploration by locally shifting waypoints to reachable neighboring locations. These strategies enrich the spatial and temporal variability of the rendered videos, improving realism and robustness.

\subsection{Tyro Tour Videos}
Tyro tours represent the simplest extension from the Gold Route.  

\noindent \textbf{Randomized Start/Goal Offsets.}  
We extend both the start and goal points by sampling reachable offsets uniformly within:
\[
d \sim \text{Uniform}(1,4).
\]
The sampled points are projected onto the navigable mesh to ensure validity.  

\noindent \textbf{Behavior Characteristics.}  
No additional exploration beyond the extended endpoints is performed; the route follows the gold path except for the offset regions.  

\noindent \textbf{Success Criterion.} 
A tour is considered successful when the ending position is within $0.25\,\text{m}$ of the sampled target endpoint.

\subsection{Curiosity Tour Videos}
Curiosity tours simulate mild exploratory behaviors driven by human-like curiosity.

\noindent \textbf{Randomized Start/Goal Offsets.}  
Offsets are sampled from a truncated Gaussian:
\[
d \sim \text{Normal}(\mu=3.6,\, \sigma=1.2,\; \text{min}=0,\; \text{max}=10).
\]

\noindent \textbf{Neighborhood Visits.}  
With probability $P$, the agent performs a forward neighborhood exploration by moving toward directions sampled from $[-90^\circ,\, +90^\circ]$ relative to its heading.  
Step lengths follow:
\[
r \sim \text{Normal}(\mu=1.5,\, \sigma=0.5,\; \text{min}=0,\; \text{max}=3).
\]

\noindent \textbf{Random Look-around.}  
Also with probability $P$, the agent performs a curiosity-driven look-around, rotating $30^\circ$ to the left and right before resuming navigation.  

\noindent \textbf{Success Criterion.}  
A tour is valid if the final position falls within $0.5\,\text{m}$ of the sampled target endpoint.

\subsection{Explorer Tour Videos}
Explorer tours introduce strong exploratory behaviors and substantially reduce trajectory overlap with the Gold Route.

\noindent \textbf{Randomized Start/Goal Offsets.}  
Offsets follow:
\[
d \sim \text{Normal}(\mu=7.5,\, \sigma=2.5,\; \text{min}=0,\; \text{max}=15).
\]

\noindent \textbf{Large-Scale Neighborhood Visits.}  
With probability $P$, neighborhood exploration is triggered in more extreme lateral directions sampled from:
\[
[-90^\circ, -30^\circ] \cup [30^\circ, 90^\circ],
\]
with step lengths:
\[
r \sim \text{Normal}(\mu=3,\, \sigma=1,\; \text{min}=0,\; \text{max}=15).
\]

\noindent \textbf{Random Look-around.}  
Similar to curiosity tours, an additional left-right $30^\circ$ look-around may be triggered with probability~$P$.  

\noindent \textbf{Success Criterion.}  
A valid explorer tour terminates within $1\,\text{m}$ of the sampled target endpoint.

\subsection{Trigger Probability $P$}
Exploration behaviors (neighborhood visits and look-arounds) are triggered with different probabilities depending on the region of the route:
\noindent \textbf{Offset regions} (random start $\rightarrow$ gold start, gold goal $\rightarrow$ random goal):  
    \[
    P = 0.02.
    \]
\noindent \textbf{Gold-route region} (gold start $\rightarrow$ gold goal):  
    \[
    P = 0.07.
    \]

This design encourages more exploration where the environment is well structured (the Gold Route region), while keeping the offset regions lightweight.

\subsection{Quality Assurance}
All tours undergo:

\noindent (1) strict connectivity validation on the navigation mesh,

\noindent (2) obstacle avoidance checks,

\noindent (3) per-frame collision detection,

\noindent (4) manual inspection for temporal consistency.

\noindent Only videos that pass all checks are included in VideoNIG.

\section{Action Warmup}
\label{sec: action_warmup}

The Action Warmup stage is designed to establish foundational visual-spatial reasoning by training the model on local spatial transformations and orientation alignment before it tackles complex, long-horizon navigation instructions. We formulate this stage through two complementary subtasks: \textit{Action-from-View} and \textit{View-from-Action}.

\subsection{Data Collection Pipeline}
To generate training samples for this stage, we programmatically interact with the Habitat simulator to collect precise (initial view, action, next view) triplets. We define a fine-grained, discrete action space consisting of 13 actions: 
\begin{itemize}
    \item \textbf{Translation}: Move forward by $\{25, 50, 75, 100\}$ cm.
    \item \textbf{Rotation}: Turn left or right by $\{15^\circ, 30^\circ, 45^\circ, 60^\circ\}$.
    \item \textbf{Stop}: Remain in the current position.
\end{itemize}
For each randomly sampled valid starting position in the environment, we first enforce a strict navigability check (e.g., ensuring a 100\,cm forward movement is collision-free). Once validated, we record the initial egocentric RGB observation. The agent then resets to the initial state and iteratively executes each of the 13 actions, capturing the corresponding subsequent observations. From this candidate pool, we systematically construct pairs for the two warmup tasks, ensuring balanced sampling and the rigorous exclusion of invalid trajectories.

\subsection{Task Formulation}
Using the collected observations, we formulate two distinct training tasks:

\noindent \textbf{Action-from-View:} Given an initial frame and a subsequent frame, the model is tasked with inferring the relative camera motion (i.e., outputting the exact action label, such as ``Move forward 100 cm'').

\noindent \textbf{View-from-Action:} Given an initial frame and a specific textual action instruction, the model must select the correct resulting frame from a candidate set. We formulate this as a multiple-choice QA task with four visual options: one ground-truth frame and three distractor frames sampled from other actions originating from the same initial state.

\subsection{GRPO Reward Formulation}
To optimize the model effectively during the Action Warmup stage using Group Relative Policy Optimization (GRPO), we carefully design task-specific reward functions. Rather than relying on a sparse binary reward, we implement a \textit{hierarchical step reward} for the Action-from-View task to capture the partial correctness of continuous spatial concepts.

\noindent \textbf{Action-from-View Reward:} The reward $R_{\text{a2v}}$ penalizes the model based on the semantic and physical distance between the predicted action and the ground truth.
\begin{itemize}
    \item \textit{Distance Hierarchy (Forward Actions):} If the ground truth is ``Move forward 100 cm'', predictions of $\{100, 75, 50, 25\}$ cm yield rewards of $\{1.0, 0.8, 0.6, 0.1\}$ respectively.
    \item \textit{Angle Hierarchy (Turn Actions):} If the ground truth is ``Turn left $30^\circ$'', predicting the exact angle yields $1.0$. Predicting adjacent angles in the same direction (e.g., $15^\circ$ or $45^\circ$) yields $0.8$, and $60^\circ$ yields $0.6$. Predicting any turn in the opposite direction (e.g., right turn) yields $0.0$.
    \item \textit{Type Mismatch:} Any confusion between fundamentally different action categories (e.g., predicting a Turn when the ground truth is Forward, or confusing Any action with Stop) strictly yields a reward of $0.0$.
\end{itemize}

\noindent \textbf{View-from-Action Reward:} Because this is formulated as an explicit multiple-choice task, we apply a straightforward binary reward $R_{\text{v2a}}$. The model receives a reward of $1.0$ if it successfully selects the correct target frame from the four candidates, and $0.0$ otherwise. 

\noindent By applying these targeted rewards, the model develops a robust prior for egocentric spatial transformations, significantly stabilizing the subsequent training phase on long, complex tour videos.

\section{Complexity Progression}
\label{sec: training_settings}
% 整体描述
% The second stage of our curriculum learning focuses on video difficulty progression, where training samples are gradually introduced according to the complexity of tour videos. 
% This stage aims to improve the model’s ability to handle long-horizon spatial reasoning and trajectory composition. To optimize the model under this curriculum, we employ two training strategies: Group Relative Policy Optimization (GRPO) and Mixed Preference Optimization (MPO), which provide complementary learning signals for navigation instruction generation.

Our curriculum learning begins with Action Warmup, which establishes a foundational spatial-orientation representation through local action-view alignment.
Building on this initialization, the second stage focuses on complexity progression, where training samples are gradually introduced according to the difficulty and exploratory behavior of the tour videos.
This progression aims to scale the model’s capacity to handle long-horizon spatial reasoning.
To optimize the model under this curriculum, we employ Group Relative Policy Optimization (GRPO) and Mixed Preference Optimization (MPO) to ensure both textual alignment and spatial-temporal consistency in navigation instruction generation.

\subsection{GRPO}
\label{sec: GRPO_reward_funcations}

\noindent\textbf{Group Relative Policy Optimization (GRPO)}~\cite{shao2024deepseekmath} is adopted to optimize instruction generation in a group-relative manner. 
A known issue of GRPO is that when all sampled outputs $\{y^i\}_{i=1}^G$ receive nearly identical rewards, the group contributes negligible gradients. 
To maintain effective batch diversity, we discard non-informative groups whose reward range falls below a threshold $\delta$ (set to 0.1).

\noindent\textbf{Objective Function.} We define the query $q$ of each sample as $(\mathcal{P}, \mathcal{V}, \mathcal{O}_0, \mathcal{G})$, which consists of the prompt, tour video, initial RGB observation, and goal description. The generated navigation instruction is denoted as $I_{\text{pred}}$. The objective function is then defined as:
\begin{equation}
\resizebox{\dimexpr\linewidth-4em}{!}{$\displaystyle
\begin{aligned}
\mathcal{J}(\theta) &= \mathbb{E}_{(q,I_{\text{gt}})\sim \mathcal{D},\, \{I_{\text{pred}}^i\}_{i=1}^G\sim \pi_{\theta_\text{old}}(\cdot\mid q)} \\
&\quad \Bigg[ \frac{1}{G} \sum_{i=1}^{G} \frac{1}{|I_{\text{pred}}^i|} \sum_{t=1}^{|I_{\text{pred}}^i|} \min \Big( r_{i,t}(\theta) \hat{A}_{i,t},\; \text{clip}\big( r_{i,t}(\theta), 1 - \varepsilon_{\text{low}}, 1 + \varepsilon_{\text{high}} \big) \hat{A}_{i,t} \Big) - \beta D_{\text{KL}}(\pi_{\theta} \| \pi_{\text{ref}}) \Bigg]
\end{aligned}
$}
\label{eq:rl_loss}
\end{equation}
% \vspace{-.5em}
{\footnotesize
\begin{equation*}
\text{s.t.}\quad \max \{R_i\}_{i=1}^G - \min \{R_i\}_{i=1}^G \ge \delta
\end{equation*}
}
% \vspace{-.3em}
where
{\footnotesize
\begin{equation*}
\hat{A}_{i,t} = \frac{R_i - \text{mean}(\{R_i\}_{i=1}^G)}{\text{std}(\{R_i\}_{i=1}^G)}, \qquad
r_{i,t}(\theta) = \frac{\pi_{\theta}(I_{\text{pred}}^{i,t} \mid q, I_{\text{pred}}^{i,<t})}{\pi_{\theta_{\text{old}}}(I_{\text{pred}}^{i,t} \mid q, I_{\text{pred}}^{i,<t})}.
\end{equation*}
}
$(q, I_{\text{gt}})$ is a (query, ground-truth instruction) pair from the training dataset $\mathcal{D}$; $\varepsilon_{\text{low}}$ and $\varepsilon_{\text{high}}$ are clipping parameters, and $G$ is the rollout group size.

\noindent \textbf{Reward functions.}
We analyze the predicted navigation instruction from three perspectives:

% \begin{itemize}[leftmargin=*,nosep]
\noindent \textbf{(1) Similarity}: We measure the similarity between the predicted instruction \(I_{\text{pred}}\) and the ground-truth instruction \(I_{\text{gt}}\) using classical metrics, including SPICE score \(f_{\text{SPICE}}(I_{\text{pred}}, I_{\text{gt}})\) and Rouge-L score \(f_{\text{Rouge-L}}(I_{\text{pred}}, I_{\text{gt}})\).

\noindent \textbf{(2) Quality}: We assess whether the predicted instruction clearly expresses each action, movement, and direction, without including extraneous information unrelated to navigation. Note that this quality score \(f_{\text{quality}}(I_{\text{pred}}\)) focuses on the clarity of expression rather than the accuracy of actions, and is evaluated by an LLM judge.

\noindent \textbf{(3) Trajectory Consistency}: We measure the consistency between the trajectory guided by the predicted instruction and the optimal trajectory recorded in the Gold Route video \(V_{\text{gt}}\). This trajectory score \(f_{\text{traj}}(V_{\text{gt}}, I_{\text{pred}})\) is evaluated by an MLLM that takes both the Gold Route video and the predicted instruction as inputs.
% \end{itemize}
The main reward component is defined as:
\begin{equation}
\textcolor{red}{
\resizebox{\dimexpr\linewidth-4em}{!}{$\displaystyle
R_{\text{main}} = \alpha f_{\text{SPICE}}(I_{\text{pred}}, I_{\text{gt}}) + \beta f_{\text{Rouge-L}}(I_{\text{pred}}, I_{\text{gt}}) + \gamma f_{\text{quality}}(I_{\text{pred}}) + \delta f_{\text{traj}}(V_{\text{gt}}, I_{\text{pred}})
$}}
\end{equation}
where \(\alpha\), \(\beta\), \(\gamma\), and \(\delta\) are the weights for each reward component.
We set \(\alpha = 0.1\), \(\beta = 0.1\), \(\gamma = 0.2\) and \(\delta = 0.6\).
\noindent To penalize overly long instructions, we define a length penalty term based on the ratio \( r = \frac{L_{\text{pred}}}{L_{\text{gt}}} \), where \( L_{\text{pred}} \) and \( L_{\text{gt}} \) are the lengths of the predicted and ground-truth instructions, respectively:

% red color

\begin{equation}
    \textcolor{red}{
    R_{\text{length}} = 
    \begin{cases}
    0, & \text{if } r < 3 \\
    -\dfrac{r - 3}{2}, & \text{if } 3 \leq r \leq 5 \\
    -1, & \text{if } r > 5
    \end{cases}}
\end{equation}
The final reward is:
\begin{equation}
    \textcolor{red}{R = R_{\text{main}} + R_{\text{length}}}
\end{equation}

\subsection{MPO.}
\label{sec: MPO_dataset}

\noindent\textbf{Mixed Preference Optimization (MPO).}\cite{wang2024enhancing} To generate semantically consistent navigation instructions from a given video $v$ and start--goal pair $(s, g)$, 
we formulate the task as conditional generation with a multimodal language model $\pi_\theta(y \mid v, s, g)$, where $y$ denotes the navigation instruction. 
The model is first initialized via supervised fine-tuning (SFT):
\begin{equation}
    \mathcal{L}_{\text{SFT}}
    =
    -\mathbb{E}_{(v,s,g,y^*)}
    \left[
    \log \pi_\theta(y^* \mid v, s, g)
    \right],
\end{equation}
where $y^*$ is the reference instruction. 
However, SFT alone does not explicitly penalize reasoning inconsistencies such as incorrect directional descriptions, entity hallucinations, or sub-path mismatches.
We therefore employ Mixed Preference Optimization (MPO), 
which decomposes instruction quality into multiple reasoning dimensions. 
For dimension $k$, we construct a preferred instruction pair $(y_k^+, y_k^-)$ and optimize:
\begin{equation}
    \mathcal{L}_{\text{pref}}^{(k)}
    =
    -\log \sigma
    \left(
    \beta
    \left(
    \log \pi_\theta(y_k^+ \mid v, s, g)
    -
    \log \pi_\theta(y_k^- \mid v, s, g)
    \right)
    \right).
\end{equation}
The mixed preference objective is

\begin{equation}
\mathcal{L}_{\text{MPO}}
=
\sum_{k=1}^{K}
\lambda_k
\mathcal{L}_{\text{pref}}^{(k)},
\end{equation}
and the final training objective integrates imitation learning, 
mixed preference optimization, and KL regularization:
\begin{equation}
    \mathcal{L}
    =
    \lambda_{\text{SFT}} \mathcal{L}_{\text{SFT}}
    +
    \mathcal{L}_{\text{MPO}}
    +
    \lambda_{\text{KL}}
    D_{\text{KL}}
    \left(
    \pi_\theta(\cdot \mid v, s, g)
    \;\|\;
    \pi_{\text{ref}}(\cdot \mid v, s, g)
    \right).
\end{equation}

\begin{table*}[t]
    \centering
    \caption{Evaluation results on VideoNIG. Each cell shows two scores: Rouge-L on top, SPICE below.}
    \label{tab:res_all_spice_suppl}
    \resizebox{\linewidth}{!}{
        \begin{tabular}{l|cc|cc|cc|cc|cc|cc|cc|cc|c}
            \toprule[1.1pt]
            Video type & \multicolumn{4}{c|}{Gold Route} & \multicolumn{4}{c|}{Tyro Tour} & \multicolumn{4}{c|}{Curiosity Tour} & \multicolumn{4}{c|}{Explorer Tour} & \multirow{3}{*}{Average} \\
            \cline{1-17}
            Source & \multicolumn{2}{c|}{R2R} & \multicolumn{2}{c|}{RxR} & \multicolumn{2}{c|}{R2R} & \multicolumn{2}{c|}{RxR} & \multicolumn{2}{c|}{R2R} & \multicolumn{2}{c|}{RxR} & \multicolumn{2}{c|}{R2R} & \multicolumn{2}{c|}{RxR} & \\
            \cline{1-17}
            Goal modal & image & text & image & text & image & text & image & text & image & text & image & text & image & text & image & text & \\
            \midrule[0.7pt]
            % 0.167	0.160	0.186	0.181	0.164	0.159	0.181	0.180	0.166	0.161	0.181	0.179	0.168	0.154	0.187	0.174	0.172
            % 0.092	0.078	0.135	0.127	0.084	0.076	0.129	0.130	0.088	0.078	0.137	0.122	0.089	0.071	0.140	0.121	0.106
            gpt-5.2-chat & \makecell{0.167\\0.092} &\makecell{0.160\\0.078} & \makecell{0.186\\0.135} & \makecell{0.181\\0.127} &\makecell{0.164\\0.084} & \makecell{0.159\\0.076} & \makecell{0.181\\0.129} & \makecell{0.180\\0.130} & \makecell{0.166\\0.088} & \makecell{0.161\\0.078} & \makecell{0.181\\0.137} & \makecell{0.179\\0.122} & \makecell{0.168\\0.089} & \makecell{0.154\\0.071} & \makecell{0.187\\0.140} & \makecell{0.174\\0.121} & \makecell{0.172\\0.106} \\
            \hline
            % 0.143	0.135	0.157	0.156	0.147	0.140	0.157	0.153	0.138	0.136	0.156	0.154	0.134	0.133	0.154	0.150	0.146
            % 0.062	0.055	0.096	0.091	0.066	0.063	0.099	0.089	0.066	0.063	0.096	0.101	0.063	0.063	0.088	0.091	0.078
            qwen3-vl-plus & \makecell{0.143\\0.062}& \makecell{0.135\\0.055} & \makecell{0.157\\0.096} & \makecell{0.156\\0.091} & \makecell{0.147\\0.066} & \makecell{0.140\\0.063} & \makecell{0.157\\0.099} & \makecell{0.153\\0.089} & \makecell{0.138\\0.066} & \makecell{0.136\\0.063} & \makecell{0.156\\0.096} & \makecell{0.154\\0.101} & \makecell{0.134\\0.063} & \makecell{0.133\\0.063} & \makecell{0.154\\0.088} & \makecell{0.150\\0.091} & \makecell{0.146\\0.078} \\
            \midrule[0.7pt]
            InternVL3-8B~\cite{zhu2025internvl3} & \makecell{0.160\\0.087} & \makecell{0.165\\0.087} & \makecell{0.173\\0.117} & \makecell{0.173\\0.112} & \makecell{0.158\\0.088} & \makecell{0.163\\0.086} & \makecell{0.174\\0.114} & \makecell{0.174\\0.113} & \makecell{0.157\\0.086} & \makecell{0.163\\0.086} & \makecell{0.173\\0.116} & \makecell{0.173\\0.109} & \makecell{0.153\\0.085} & \makecell{0.162\\0.085} & \makecell{0.172\\0.114} & \makecell{0.174\\0.111} & \makecell{0.167\\0.100} \\
            \hline
            InternVL3.5-4B~\cite{wang2025internvl35} & \makecell{0.191\\0.105} & \makecell{0.174\\0.097} & \makecell{0.173\\0.116} & \makecell{0.174\\0.115} & \makecell{0.189\\0.101} & \makecell{0.170\\0.090} & \makecell{0.174\\0.116} & \makecell{0.173\\0.114} & \makecell{0.188\\0.101} & \makecell{0.169\\0.091} & \makecell{0.174\\0.116} & \makecell{0.172\\0.112} & \makecell{0.187\\0.101} & \makecell{0.168\\0.089} & \makecell{0.172\\0.116} & \makecell{0.173\\0.111} & \makecell{0.176\\0.106} \\
            \hline
            InternVL3.5-8B~\cite{wang2025internvl35} & \makecell{0.217\\0.109} & \makecell{0.212\\0.105} & \makecell{0.174\\0.112} & \makecell{0.176\\0.114} & \makecell{0.213\\0.109} & \makecell{0.210\\0.102} & \makecell{0.174\\0.113} & \makecell{0.177\\0.114} & \makecell{0.212\\0.106} & \makecell{0.209\\0.103} & \makecell{0.174\\0.113} & \makecell{0.176\\0.112} & \makecell{0.209\\0.108} & \makecell{0.206\\0.100} & \makecell{0.174\\0.111} & \makecell{0.176\\0.113} & \makecell{0.193\\0.109} \\
            \midrule[0.7pt]
            Qwen2.5-VL-7B~\cite{bai2025qwen2_5} & \makecell{0.195\\0.109} & \makecell{0.182\\0.094} & \makecell{0.164\\0.101} & \makecell{0.171\\0.102} & \makecell{0.199\\0.106} & \makecell{0.178\\0.095} & \makecell{0.166\\0.097} & \makecell{0.175\\0.104} & \makecell{0.198\\0.106} & \makecell{0.178\\0.093} & \makecell{0.166\\0.099} & \makecell{0.173\\0.102} & \makecell{0.195\\0.105} & \makecell{0.177\\0.094} & \makecell{0.166\\0.100} & \makecell{0.175\\0.105} & \makecell{0.179\\0.101} \\
            \hline
            Qwen3-VL-4B~\cite{bai2025qwen3vl} & \makecell{0.224\\0.114} & \makecell{0.201\\0.097} & \makecell{0.172\\0.131} & \makecell{0.181\\0.118} & \makecell{0.169\\0.065} & \makecell{0.097\\0.053} & \makecell{0.187\\0.095} & \makecell{0.195\\0.136} & \makecell{0.243\\0.107} & \makecell{0.201\\0.091} & \makecell{0.192\\0.132} & \makecell{0.180\\0.116} & \makecell{0.217\\0.110} & \makecell{0.196\\0.091} & \makecell{0.201\\0.141} & \makecell{0.178\\0.113} & \makecell{0.190\\0.107} \\
            \hline
            Qwen3-VL-8B~\cite{bai2025qwen3vl} & \makecell{0.213\\0.121} & \makecell{0.225\\0.098} & \makecell{0.175\\0.111} & \makecell{0.176\\0.107} & \makecell{0.212\\0.117} & \makecell{0.201\\0.101} & \makecell{0.176\\0.111} & \makecell{0.175\\0.104} & \makecell{0.210\\0.117} & \makecell{0.199\\0.100} & \makecell{0.175\\0.109} & \makecell{0.174\\0.103} & \makecell{0.212\\0.116} & \makecell{0.199\\0.096} & \makecell{0.173\\0.107} & \makecell{0.174\\0.102} & \makecell{0.192\\0.107} \\
            \midrule[0.7pt]
            Qwen3-VL-4B (SFT) & \makecell{0.277\\0.156} & \makecell{0.282\\0.154} & \makecell{0.201\\0.157} & \makecell{0.202\\0.156} & \makecell{0.267\\0.148} & \makecell{0.269\\0.151} & \makecell{0.202\\0.158} & \makecell{0.201\\0.155} & \makecell{0.254\\0.147} & \makecell{0.257\\0.143} & \makecell{0.200\\0.155} & \makecell{0.200\\0.155} & \makecell{0.240\\0.134} & \makecell{0.243\\0.138} & \makecell{0.199\\0.154} & \makecell{0.197\\0.151} & \makecell{0.231\\0.151} \\
            \hline
            Qwen3-VL-8B (SFT) & \makecell{0.283\\0.156} & \makecell{0.284\\0.159} & \makecell{0.197\\0.151} & \makecell{0.197\\0.152} & \makecell{0.274\\0.152} & \makecell{0.278\\0.156} & \makecell{0.195\\0.147} & \makecell{0.196\\0.148} & \makecell{0.267\\0.149} & \makecell{0.271\\0.146} & \makecell{0.195\\0.148} & \makecell{0.193\\0.147} & \makecell{0.258\\0.143} & \makecell{0.259\\0.139} & \makecell{0.193\\0.146} & \makecell{0.191\\0.143} & \makecell{0.233\\0.149} \\
            \hline
            Qwen3-VL-8B (GRPO) & \makecell{0.301\\0.196} & \makecell{0.299\\0.195} & \makecell{0.209\\0.194} & \makecell{0.209\\0.190} & \makecell{0.301\\0.193} & \makecell{0.299\\0.193} & \makecell{0.211\\0.195} & \makecell{0.210\\0.189} & \makecell{0.298\\0.194} & \makecell{0.297\\0.190} & \makecell{0.211\\0.196} & \makecell{0.210\\0.191} & \makecell{0.296\\0.190} & \makecell{0.294\\0.185} & \makecell{0.211\\0.189} & \makecell{0.209\\0.186} & \makecell{\textbf{0.254}\\\textbf{0.192}} \\

            \hline
            Qwen3-VL-8B (MPO) & \makecell{0.348\\0.196} & \makecell{0.340\\0.187} & \makecell{0.186\\0.126} & \makecell{0.182\\0.121} & \makecell{0.334\\0.186} & \makecell{0.315\\0.170} & \makecell{0.191\\0.131} & \makecell{0.187\\0.133} & \makecell{0.296\\0.173} & \makecell{0.270\\0.151} & \makecell{0.179\\0.133} & \makecell{0.171\\0.125} & \makecell{0.266\\0.161} & \makecell{0.226\\0.136} & \makecell{0.166\\0.126} & \makecell{0.156\\0.123} & \makecell{0.238\\0.149} \\
            \hline
            Qwen3-VL-8B (MPO-lora) & \makecell{0.331\\0.190} & \makecell{0.335\\0.192} & \makecell{0.192\\0.149} & \makecell{0.182\\0.129} & \makecell{0.336\\0.195} & \makecell{0.338\\0.191} & \makecell{0.190\\0.144} & \makecell{0.185\\0.128} & \makecell{0.334\\0.189} & \makecell{0.327\\0.186} & \makecell{0.193\\0.147} & \makecell{0.183\\0.128} & \makecell{0.326\\0.185} & \makecell{0.306\\0.168} & \makecell{0.184\\0.136} & \makecell{0.182\\0.126} & \makecell{0.258\\0.161} \\
            \bottomrule[1.1pt]
        \end{tabular}
    }
\end{table*}

\noindent \textbf{Dataset.}
During MPO training, preference pairs are constructed to guide optimization. 
Specifically, we design five types of negative samples as rejected instructions (\cref{fig:neg_choice}).
We use \texttt{qwen3-vl-plus} to generate these instructions, producing diverse distractor descriptions conditioned on the original trajectory.
For each negative sample type, we generate three variants. 
This results in a pool of $5 \times 3 = 15$ candidate rejected instructions. 
During training, one instruction is randomly sampled from this pool to construct the preference pair.

\begin{table}[t]
    \centering
    \caption{Evaluation results on VideoNIG. Each cell shows choice-evaluation results: Multiple-choice on top and Orthogonal-choice on the bottom. Values are percentages.}
    \label{tab:res_all_choices_suppl}
    \resizebox{\linewidth}{!}{
        \begin{tabular}{l|cc|cc|cc|cc|cc|cc|cc|cc|c}
            \toprule[1.1pt]
            Video type & \multicolumn{4}{c|}{Gold Route} & \multicolumn{4}{c|}{Tyro Tour} & \multicolumn{4}{c|}{Curiosity Tour} & \multicolumn{4}{c|}{Explorer Tour} & \multirow{3}{*}{Average} \\
            \cline{1-17}
            Source & \multicolumn{2}{c|}{R2R} & \multicolumn{2}{c|}{RxR} & \multicolumn{2}{c|}{R2R} & \multicolumn{2}{c|}{RxR} & \multicolumn{2}{c|}{R2R} & \multicolumn{2}{c|}{RxR} & \multicolumn{2}{c|}{R2R} & \multicolumn{2}{c|}{RxR} & \\
            \cline{1-17}
            Goal modal & image & text & image & text & image & text & image & text & image & text & image & text & image & text & image & text & \\
            \midrule[0.7pt]
            gpt-5.2-chat & \makecell{63.00\\64.00} &\makecell{53.00\\65.00} & \makecell{56.00\\52.00} & \makecell{52.00\\54.00} &\makecell{51.00\\58.00} & \makecell{55.00\\53.00} & \makecell{60.00\\57.00} & \makecell{49.00\\59.00} & \makecell{54.00\\58.00} & \makecell{50.00\\58.00} & \makecell{52.00\\44.00} & \makecell{59.00\\52.00} & \makecell{55.00\\54.00} & \makecell{61.00\\52.00} &\makecell{56.00\\57.00} & \makecell{54.00\\58.00} & \makecell{55.00\\56.13} \\
            \hline
            qwen3-vl-plus & \makecell{63.00\\59.00}& \makecell{63.00\\66.00} & \makecell{60.00\\52.00} & \makecell{58.00\\52.00} & \makecell{58.00\\60.00} & \makecell{58.00\\62.00} & \makecell{65.00\\56.00} & \makecell{65.00\\53.00} & \makecell{61.00\\60.00} & \makecell{56.00\\67.00} & \makecell{63.00\\55.00} & \makecell{60.00\\54.00} & \makecell{60.00\\55.00} & \makecell{60.00\\52.00} & \makecell{54.00\\48.00} & \makecell{62.00\\46.00} & \makecell{60.38\\56.07} \\
            \midrule[0.7pt]
            InternVL3-8B~\cite{zhu2025internvl3} & \makecell{43.97\\47.77} & \makecell{43.13\\47.10} & \makecell{43.62\\44.79} & \makecell{42.98\\43.10} & \makecell{42.16\\45.58} & \makecell{43.98\\45.05} & \makecell{39.56\\41.02} & \makecell{39.49\\44.73} & \makecell{41.27\\--46.86} & \makecell{41.18\\45.48} & \makecell{39.64\\41.26} & \makecell{40.53\\43.85} & \makecell{39.90\\45.06} & \makecell{40.35\\45.23} & \makecell{41.35\\41.80} & \makecell{40.59\\41.63} & \makecell{41.43\\44.11} \\
            \hline
            InternVL3.5-4B~\cite{wang2025internvl35} & \makecell{43.97\\47.77} & \makecell{43.13\\47.10} & \makecell{43.62\\44.79} & \makecell{42.98\\43.10} & \makecell{42.16\\45.58} & \makecell{43.98\\45.05} & \makecell{39.56\\41.02} & \makecell{39.49\\44.73} & \makecell{41.27\\46.86} & \makecell{41.18\\45.48} & \makecell{39.64\\41.26} & \makecell{40.53\\43.85} & \makecell{39.90\\45.06} & \makecell{40.35\\45.23} & \makecell{41.35\\41.80} & \makecell{40.59\\41.63} & \makecell{41.48\\44.39} \\
            \hline
            InternVL3.5-8B~\cite{wang2025internvl35} & \makecell{44.55\\49.54} & \makecell{49.40\\52.77} & \makecell{44.32\\50.78} & \makecell{44.89\\48.49} & \makecell{43.85\\48.35} & \makecell{45.19\\47.19} & \makecell{41.57\\45.42} & \makecell{44.76\\45.29} & \makecell{44.82\\49.51} & \makecell{45.03\\49.59} & \makecell{42.74\\45.78} & \makecell{44.24\\46.03} & \makecell{44.02\\48.55} & \makecell{46.33\\48.85} & \makecell{43.04\\45.36} & \makecell{42.88\\45.26} & \makecell{44.48\\47.92} \\
            \midrule[0.7pt]
            Qwen3-VL-4B~\cite{bai2025qwen3vl} & \makecell{48.19\\52.21} & \makecell{48.50\\53.50} & \makecell{41.48\\50.81} & \makecell{42.07\\49.48} & \makecell{45.15\\51.05} & \makecell{46.65\\51.92} & \makecell{42.27\\48.23} & \makecell{43.45\\50.00} & \makecell{49.10\\50.00} & \makecell{47.87\\50.09} & \makecell{41.39\\46.63} & \makecell{43.83\\47.58} & \makecell{45.38\\52.33} & \makecell{46.64\\51.20} & \makecell{44.45\\45.75} & \makecell{41.59\\48.01} & \makecell{44.88\\49.92} \\
            \hline
            Qwen3-VL-8B~\cite{bai2025qwen3vl} & \makecell{54.95\\52.70} & \makecell{54.31\\55.24} & \makecell{47.45\\53.50} & \makecell{49.04\\54.42} & \makecell{52.86\\54.89} & \makecell{56.33\\51.34} & \makecell{47.85\\51.52} & \makecell{48.36\\50.92} & \makecell{55.44\\53.85} & \makecell{57.04\\52.81} & \makecell{46.13\\50.99} & \makecell{48.04\\49.41} & \makecell{52.50\\52.25} & \makecell{54.50\\52.04} & \makecell{46.66\\49.55} & \makecell{45.45\\48.72} & \makecell{51.06\\52.13} \\
            \hline
            Qwen3-VL-8B (SFT) & \makecell{58.00\\57.46} & \makecell{59.03\\58.28} & \makecell{44.42\\55.84} & \makecell{43.06\\53.42} & \makecell{60.82\\55.40} & \makecell{56.57\\58.18} & \makecell{42.78\\51.96} & \makecell{43.30\\50.92} & \makecell{56.98\\54.08} & \makecell{59.02\\55.25} & \makecell{41.50\\49.48} & \makecell{42.23\\47.55} & \makecell{53.49\\53.96} & \makecell{52.85\\50.83} & \makecell{42.16\\48.84} & \makecell{40.62\\48.31} & \makecell{49.80\\53.11} \\
            \midrule[0.7pt]
            Qwen3-VL-8B (GRPO) & \makecell{55.86\\59.50} & \makecell{56.29\\59.54} & \makecell{48.65\\54.78} & \makecell{49.35\\53.33} & \makecell{55.28\\56.57} & \makecell{54.69\\57.00} & \makecell{45.94\\52.07} & \makecell{47.73\\53.15} & \makecell{54.78\\57.10} & \makecell{56.94\\56.65} & \makecell{47.53\\52.20} & \makecell{46.58\\51.14} & \makecell{52.24\\56.81} & \makecell{53.54\\54.82} & \makecell{47.34\\48.61} & \makecell{44.65\\48.75} & \makecell{51.09\\54.50} \\
            \hline
            Qwen3-VL-8B (MPO) & \makecell{62.63\\68.05} & \makecell{65.54\\68.42} & \makecell{56.13\\59.13} & \makecell{56.42\\59.38} & \makecell{61.27\\66.59} & \makecell{61.07\\65.26} & \makecell{52.39\\59.25} & \makecell{51.99\\57.57} & \makecell{58.10\\59.07} & \makecell{55.23\\57.38} & \makecell{51.19\\56.15} & \makecell{49.61\\52.71} & \makecell{54.50\\56.69} & \makecell{49.62\\51.94} & \makecell{47.22\\50.86} & \makecell{47.04\\48.33} & \makecell{55.00\\58.55} \\
            \hline
            Qwen3-VL-8B (MPO-lora) & \makecell{65.56\\68.74} & \makecell{67.44\\70.66} & \makecell{62.14\\63.52} & \makecell{58.06\\62.38} & \makecell{64.80\\67.99} & \makecell{64.90\\67.21} & \makecell{59.11\\60.55} & \makecell{59.93\\58.78} & \makecell{63.51\\65.77} & \makecell{61.90\\62.81} & \makecell{58.91\\58.34} & \makecell{59.46\\57.26} & \makecell{59.47\\62.64} & \makecell{54.31\\57.43} & \makecell{55.98\\52.86} & \makecell{54.48\\54.55} & \makecell{\textbf{60.62}\\\textbf{61.97}} \\
            \bottomrule[1.1pt]
        \end{tabular}
    }
\end{table}

\begin{table*}[t]
    \centering
    \caption{Navigation execution results on R2R and RxR with different instructions. System2 denotes InternVLA-N1 + ShortestPathFollower, and Dual System denotes InternVLA-N1 DualVLN.}
    \label{tab:nav_exec_suppl}
    \resizebox{0.9\linewidth}{!}{%
    \scriptsize
    \setlength{\tabcolsep}{1.5pt}%
    \renewcommand{\arraystretch}{0.8}%
    \begin{tabular}{l|l|l|cccc|ccccc}
        \toprule[0.8pt]
        \multirow{2}{*}{Video Type} & \multirow{2}{*}{Instruction Type} & \multirow{2}{*}{Model} & \multicolumn{4}{c|}{R2R} & \multicolumn{5}{c}{RxR} \\
        \cmidrule(lr){4-7} \cmidrule(lr){8-12}
        & & & SR $\uparrow$ & SPL $\uparrow$ & OS $\uparrow$ & NE $\downarrow$ & SR $\uparrow$ & SPL $\uparrow$ & OS $\uparrow$ & NE $\downarrow$ & nDTW $\uparrow$ \\
        \midrule[0.7pt]
        \multirow{8}{*}{Gold Route} & \multirow{2}{*}{Ground Truth} & System2 & 58.5 & 53.8 & 65.9 & 4.70 & 55.0 & 47.2 & 64.0 & 5.67 & 66.1 \\
        & & Dual System & \textbf{63.4} & \textbf{57.8} & 69.0 & \textbf{4.29} & \textbf{59.4} & \textbf{50.4} & \textbf{68.2} & \textbf{4.73} & \textbf{69.3} \\
        \cmidrule(lr){2-12}
        & \multirow{2}{*}{Qwen3-VL-8B} & System2 & 40.3 & 34.6 & 50.5 & 6.61 & 29.2 & 23.5 & 40.0 & 8.33 & 48.3 \\
        & & Dual System & 41.6 & 35.4 & 52.4 & 6.21 & 29.2 & 23.5 & 40.0 & 8.33 & 48.3 \\
        \cmidrule(lr){2-12}
        & \multirow{2}{*}{Qwen3-VL-8B(GRPO)} & System2 & 40.1 & 32.2 & 60.4 & 6.57 & 31.4 & 24.6 & 47.8 & 8.43 & 47.3 \\
        & & Dual System & 41.9 & 32.8 & 62.0 & 6.43 & 30.9 & 23.9 & 47.1 & 8.29 & 47.8 \\
        \cmidrule(lr){2-12}
        & \multirow{2}{*}{Qwen3-VL-8B(MPO)} & System2 & 56.1 & 49.8 & 66.7 & 4.84 & 37.5 & 31.6 & 49.7 & 7.23 & 55.3 \\
        & & Dual System & 60.3 & 52.7 & \textbf{69.7} & 4.33 & 40.7 & 33.3 & 51.2 & 6.83 & 56.4 \\
        \cmidrule(lr){2-12}
        & \multirow{2}{*}{Qwen3-VL-8B(MPO-lora)} & System2 & 53.6 & 47.1 & 64.6 & 4.99 & 35.8 & 29.8 & 49.8 & 7.40 & 54.4 \\
        & & Dual System & 55.6 & 49.1 & 65.5 & 4.79 & 38.7 & 31.6 & 51.0 & 7.01 & 56.2 \\
        \midrule[0.7pt]
        \multirow{4}{*}{Tyro Tour} & \multirow{2}{*}{Qwen3-VL-8B} & System2 & 31.0 & 29.2 & 39.0 & 6.42 & 20.4 & 18.7 & 26.6 & 8.87 & 45.6 \\
        & & Dual System & 33.9 & 30.4 & 36.7 & 6.26 & 20.8 & 21.5 & 28.7 & 8.93 & 46.9 \\
        \cmidrule(lr){2-12}
        & \multirow{2}{*}{Qwen3-VL-8B(MPO)} & System2 & 33.2 & 30.5 & \textbf{40.5} & 6.04 & 21.8 & 18.8 & 28.7 & 8.26 & 47.9 \\
        & & Dual System & \textbf{34.3} & \textbf{32.4} & 40.3 & \textbf{6.01} & \textbf{23.4} & \textbf{21.8} & \textbf{29.3} & \textbf{7.95} & \textbf{47.2} \\
        \cmidrule(lr){2-12}
        & \multirow{2}{*}{Qwen3-VL-8B(GRPO)} & System2 & 31.2 & 29.2 & 39.4 & 6.57 & 21.4 & 18.6 & 27.8 & 8.43 & 47.3 \\
        & & Dual System & 31.9 & 30.8 & 39.6 & 6.53 & 20.9 & 18.9 & 27.1 & 8.42 & 47.1 \\
        \cmidrule(lr){2-12}
        & \multirow{2}{*}{Qwen3-VL-8B(MPO-lora)} & System2 & 20.7 & 18.9 & 28.4 & 8.32 & 10.6 & 8.77 & 23.5 & 9.84 & 40.9 \\
        & & Dual System & 16.9 & 15.5 & 25.9 & 8.52 & 13.4 & 10.7 & 24.7 & 9.63 & 42.0 \\
        \midrule[0.7pt]
        \multirow{4}{*}{Curiosity Tour} & \multirow{2}{*}{Qwen3-VL-8B} & System2 & 21.0 & 19.2 & 29.0 & 8.22 & 10.4 & 8.77 & 23.9 & 9.87 & 40.6 \\
        & & Dual System & 16.9 & 15.4 & 26.7 & 8.26 & 12.8 & 10.5 & 24.7 & 9.63 & 41.9 \\
        \cmidrule(lr){2-12}
        & \multirow{2}{*}{Qwen3-VL-8B(MPO)} & System2 & \textbf{21.2} & \textbf{19.5} & \textbf{29.9} & \textbf{8.04} & 10.8 & 8.81 & \textbf{24.7} & 9.25 & 40.9 \\
        & & Dual System & 18.2 & 16.4 & 27.3 & 8.21 & \textbf{13.4} & \textbf{11.2} & 21.3 & \textbf{8.55} & \textbf{41.9} \\
        \cmidrule(lr){2-12}
        & \multirow{2}{*}{Qwen3-VL-8B(GRPO)} & System2 & 20.2 & 18.4 & 28.7 & 8.24 & 10.7 & 9.04 & 24.1 & 9.85 & 40.9 \\
        & & Dual System & 18.6 & 17.1 & 28.4 & 8.26 & 11.2 & 9.12 & 24.6 & 9.83 & 41.1 \\
        \cmidrule(lr){2-12}
        & \multirow{2}{*}{Qwen3-VL-8B(MPO-lora)} & System2 & 20.1 & 18.5 & 27.9 & 8.24 & 9.86 & 8.31 & 23.6 & 9.90 & 40.5 \\
        & & Dual System & 20.1 & 18.4 & 28.3 & 8.27 & 12.2 & 9.78 & 24.5 & 9.63 & 41.8 \\
        \midrule[0.7pt]
        \multirow{4}{*}{Explorer Tour} & \multirow{2}{*}{Qwen3-VL-8B} & System2 & 18.7 & 17.8 & 25.3 & 8.43 & 8.21 & 8.42 & 23.0 & 9.98 & 38.0 \\
        & & Dual System & 19.3 & 18.2 & 25.9 & \textbf{8.17} & 10.2 & 10.7 & 21.9 & 8.61 & 40.1 \\
        \cmidrule(lr){2-12}
        & \multirow{2}{*}{Qwen3-VL-8B(MPO)} & System2 & \textbf{20.4} & 18.1 & \textbf{25.9} & 8.34 & 10.3 & 8.51 & \textbf{24.2} & 9.75 & 39.2 \\
        & & Dual System & 20.2 & \textbf{18.3} & 25.3 & 8.23 & \textbf{11.4} & \textbf{10.8} & 22.0 & \textbf{8.58} & \textbf{41.2} \\
        \cmidrule(lr){2-12}
        & \multirow{2}{*}{Qwen3-VL-8B(GRPO)} & System2 & 19.1 & 17.5 & 27.6 & 8.23 & 10.6 & 8.31 & 23.1 & 9.84 & 41.0 \\
        & & Dual System & 19.9 & 17.9 & 27.9 & 8.24 & 10.6 & 8.94 & 23.5 & 9.85 & 41.2 \\
        \cmidrule(lr){2-12}
        & \multirow{2}{*}{Qwen3-VL-8B(MPO-lora)} & System2 & 21.7 & 19.8 & 28.8 & 8.23 & 10.4 & 8.79 & 23.9 & 9.73 & 40.9 \\
        & & Dual System & 18.8 & 17.0 & 27.6 & 8.31 & 12.7 & 10.4 & 24.3 & 9.46 & 41.9 \\
        \bottomrule[0.8pt]
    \end{tabular}%
    }
\end{table*}

\section{Experimental Results}
\label{sec:experimental_results}

This section provides supplementary results to complement the main experiments and instruction navigation execution evaluations presented in the primary manuscript. 
In particular, we include additional evaluations on both proprietary and open-source multimodal models to provide a more comprehensive analysis of the proposed VideoNIG task.

\subsection{Main Results}
For the VideoNIG task, we evaluate a diverse set of both open-source and proprietary multimodal models. 
The proprietary benchmarks include \texttt{gpt-5.2-chat} and \texttt{qwen3-vl-plus}, both evaluated via API access. 
Following the environmental settings of R2R and RxR, we construct 16 distinct evaluation configurations and evaluate on a representative subset of 100 randomly sampled instances for each configuration.
For open-source models, we assess the InternVL and Qwen3 series at the 4B and 8B parameter scales. 
The evaluation encompasses text-similarity metrics and choice evaluations, specifically Multiple-choice and Orthogonal-choice. Detailed results for text similarity are reported in \cref{tab:res_all_spice_suppl}, while the results for choice-based diagnostics are shown in \cref{tab:res_all_choices_suppl}.

\noindent \textbf{Text Similarity Metrics.}
We quantify the linguistic overlap between predicted and ground-truth instructions using Rouge-L and SPICE. 
As shown in \cref{tab:res_all_spice_suppl}, GRPO-trained models generally achieve the highest text-similarity scores, reflecting superior alignment with the reference instruction's surface-level phrasing. 
This is expected, as the GRPO reward function explicitly optimizes these metrics during training.
In contrast, proprietary models tend to yield lower similarity scores despite strong reasoning capabilities.
This discrepancy is primarily attributed to differences in linguistic style and phrasing relative to the ground-truth annotations rather than a lack of spatial understanding.
This observation further justifies the need for choice evaluations, namely Multiple-choice and Orthogonal-choice, to assess spatial-temporal reasoning in navigation instruction generation beyond simple textual overlap.

\noindent \textbf{Choice Evaluation.}
We evaluate model performance using Multiple-choice and Orthogonal-choice accuracy, as shown in \cref{tab:res_all_choices_suppl}.
A clear performance trend emerges: MPO-trained models achieve the highest accuracy, followed by proprietary models, while other open-source baselines generally underperform.
Specifically, the superior performance of MPO-trained models in Multiple-choice settings demonstrates the effectiveness of our two-stage curriculum framework and preference-based optimization in sharpening the model's discriminative power against diverse negative distractors. 
In Orthogonal-choice evaluations, the MPO-trained models also excel, indicating a reduced reliance on superficial textual patterns and an improved focus on core directional and entity cues. 
Despite the general-purpose strengths of proprietary models, our task-specific fine-tuned models outperform them in these structured assessments, suggesting that careful spatial-action alignment can surpass even proprietary models in specialized navigation tasks.

\begin{figure}[t]
    \centering
    \includegraphics[width=1\linewidth]{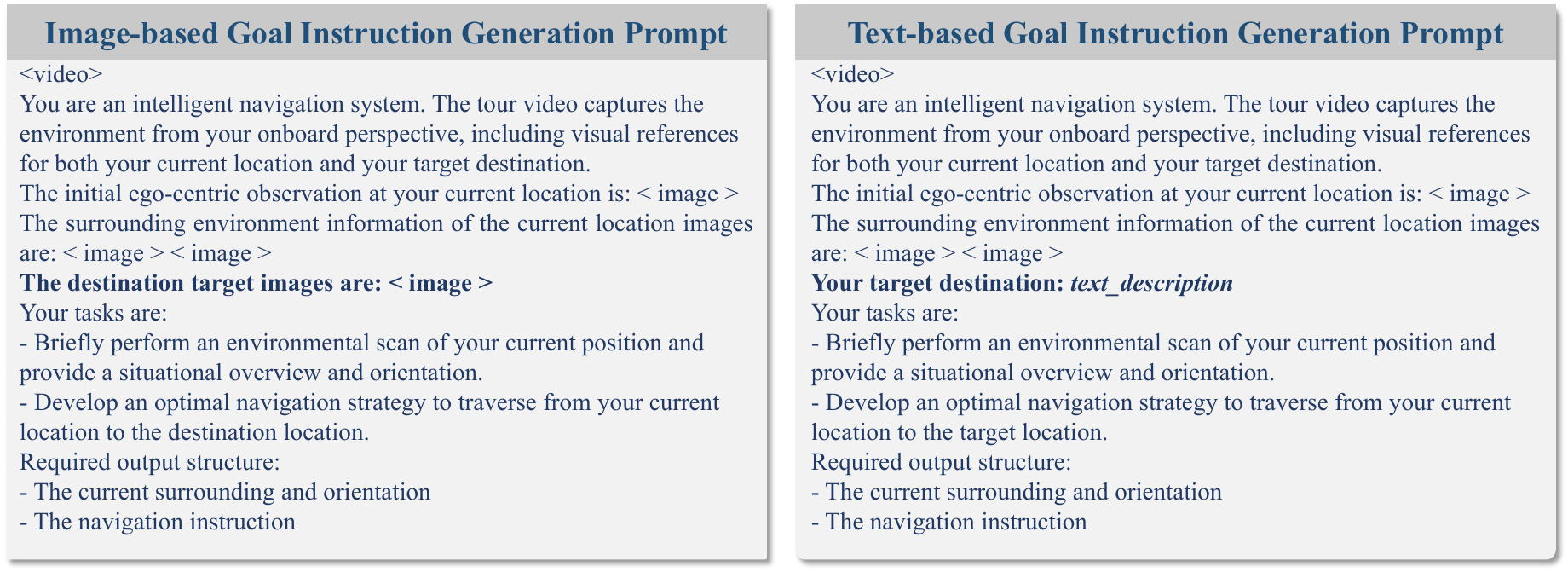}
    \caption{
        The prompt templates support goal specification via either image-based goal observations or textual descriptions in the VideoNIG task.
    }
    \label{fig:prompt}
\end{figure}

\begin{figure}[t]
    \centering
    \includegraphics[width=1\linewidth]{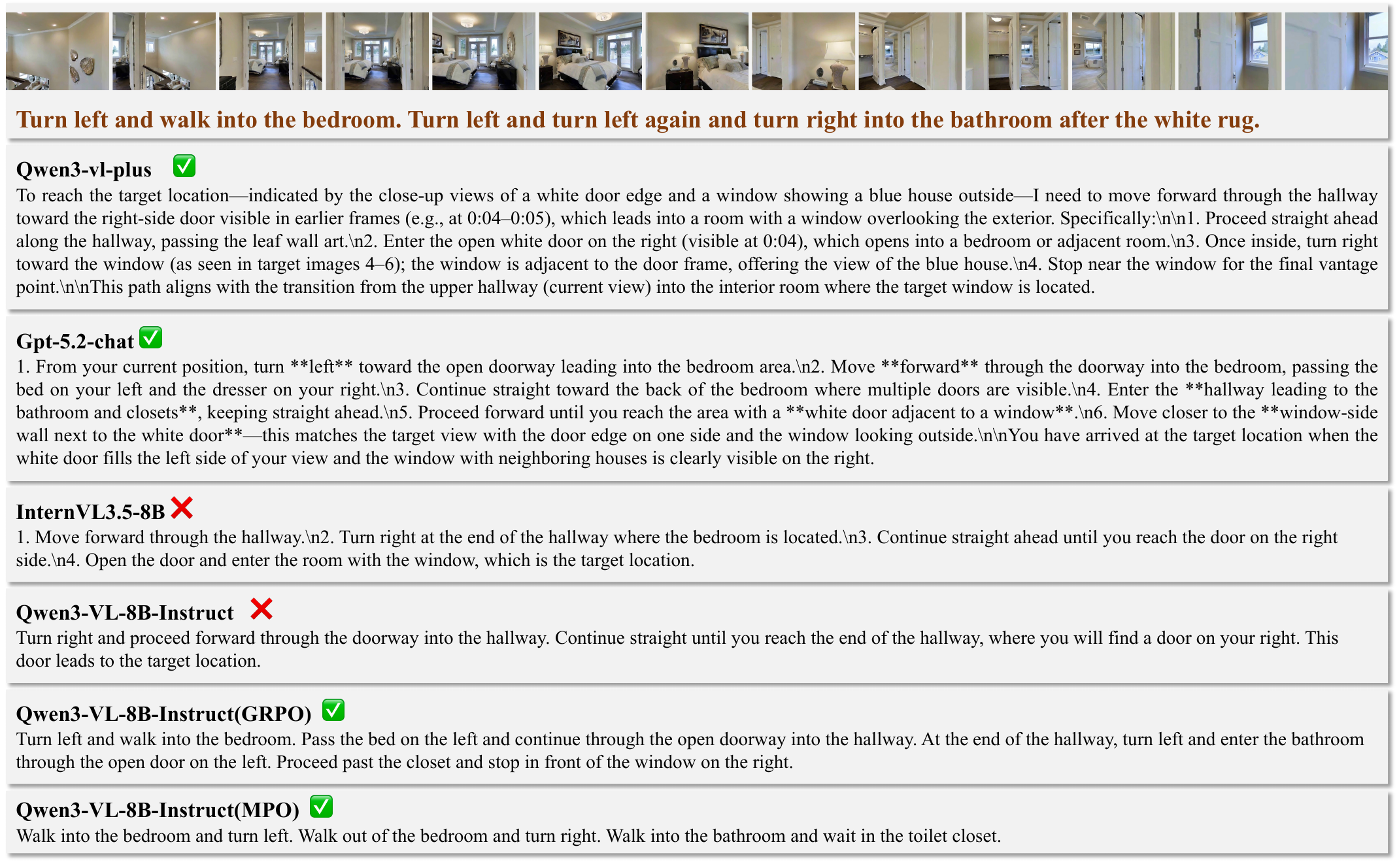}
    \caption{
        Qualitative comparison of navigation instructions generated by proprietary models, open-source models, and our fine-tuned models.
    }
    \label{fig:different_model_results}
\end{figure}

\subsection{Navigation Execution.}
% 从主实验的结果来看基于MPO训练后的LOra的模型效果会更好，这里增加了基于Lora的导航执行结果，但是从执行的结果我们可以看到相比MPO的效果会差很多，这也说明了lora的微调对齐训练可能使模型模仿输出，而并非理解视频空间。基于强化学习的结果我们可以看到，相比之下效果也会比较差，这个一方面原因是生成的指令风格可能会和Ground Truth不太相似，加上dualVLN模型的训练是基于Groundtruth训练，可能会存在一些过拟合的情况，致使在我们的数据上的性能会相对差一些。
To investigate the functional utility of the generated instructions, we conduct downstream navigation execution tests within the Habitat simulator, which provides high-fidelity continuous 3D environments.
Following the setup in our main experiments, we employ InternVLA-N1 as the navigation follower under both System-2 and Dual-System configurations.

\noindent As established in the main results, the model optimized with MPO-lora achieves superior performance across diagnostic metrics.
We additionally report the execution results for models fine-tuned using LoRA, as shown in \cref{tab:nav_exec_suppl}.
Notably, we observe that LoRA-based supervised models exhibit significantly lower navigation success rates compared to those using full-parameter fine-tuning.
This observation suggests that LoRA may lead to an imitation bias. 
In this scenario, the model primarily mimics the linguistic style of the training data instead of internalizing the underlying spatial-temporal structure. 
In contrast, reinforcement-based optimization enforces strict consistency between generated instructions and navigation logic. 
This approach effectively mitigates style mimicry and results in superior spatial grounding.

\noindent We also note that the execution performance of GRPO-generated instructions remains below that of ground-truth (GT) instructions when evaluated using the DualVLN agent. 
We attribute this remaining gap to a distributional shift in linguistic style. 
Since the DualVLN agent is pre-trained exclusively on GT instructions, it inherently develops a distributional bias toward specific human phrasing and syntactic patterns. 
Consequently, even when model-generated instructions are spatially accurate, subtle stylistic departures can degrade agent execution performance.

\section{Prompt Template}
In the VideoNIG task, the goal location can be specified via either image-based goal observations or textual descriptions. 
\cref{fig:prompt} illustrates the specific multimodal prompt templates employed for these two settings, ensuring consistent task specification across different model architectures.

\section{Qualitative Comparison of Different Models}
% 本章节我会介绍一开源模型和闭源模型的输出结果展示，我们可以看到闭源模型的输出和开源模型的输出是否正确，以及他们的输出格式，相比之下使用我们的方法训练之后输出指令格式会更加对齐Ground Truth风格，减少冗余话语的输出。

In this section, we present qualitative examples of navigation instructions generated by both open-source and proprietary models, as shown in \cref{fig:different_model_results}.
By comparing these instructions, we analyze the correctness and formatting styles of the generated instructions. 
The model trained with our method produces instructions that better align with the ground-truth style while reducing redundant expressions.

\end{document}